\documentclass{article}

\usepackage{arxiv}

\usepackage[utf8]{inputenc} 
\usepackage[T1]{fontenc}    
\usepackage{hyperref}       
\usepackage{url}            
\usepackage{booktabs}       
\usepackage{amsfonts}       
\usepackage{nicefrac}       
\usepackage{microtype}      
\usepackage{lipsum}		
\usepackage{graphicx}
\usepackage{natbib}
\usepackage{doi}

\usepackage[textsize=tiny]{todonotes}
\usepackage{listings}
\usepackage{caption}
\usepackage{subcaption}
\usepackage{array}
\usepackage{multirow}
\usepackage{makecell}
\usepackage{adjustbox}
\usepackage{enumitem}
\usepackage{amsmath} 

\title{Advancing vision-language models in front-end development via data synthesis}

\author{Tong Ge$^1$\thanks{These authors contributed equally to this work}, Yashu Liu$^{1*}$\thanks{Corresponding author: liuyashu@gmail.com}, Jieping Ye$^2$, Tianyi Li$^1$, Chao Wang$^1$\\
$^1$KE Holdings Inc. \\
$^2$Alibaba Group \\ 
}

\hypersetup{
pdftitle={Advancing vision-language models in front-end development via data synthesis}, 
}

\begin{document}
\maketitle

\begin{abstract}
	\section{Abstract}

Modern front-end (FE) development, especially when leveraging the unique features of frameworks like React and Vue, presents distinctive challenges. 
These include managing modular architectures, ensuring synchronization between data and visual outputs for declarative rendering, and adapting reusable components to various scenarios. 
Such complexities make it particularly difficult for state-of-the-art large vision-language models (VLMs) to generate accurate and functional code directly from design images.
To address these challenges, we propose a reflective agentic workflow that synthesizes high-quality image-text data to capture the diverse characteristics of FE development.
This workflow automates the extraction of self-contained\footnote{A \textbf{self-contained} code snippet is one that encapsulates all necessary logic, styling, and dependencies, ensuring it functions independently without requiring external imports or context.} code snippets from real-world projects, renders the corresponding visual outputs, and generates detailed descriptions that link design elements to functional code. 
To further expand the scope and utility of the synthesis, we introduce three data synthesis strategies: Evolution-based synthesis, which enables scalable and diverse dataset expansion; Waterfall-Model-based synthesis, which generates logically coherent code derived from system requirements; and Additive Development synthesis, which iteratively increases the complexity of human-authored components. 
We build a large vision-language model, Flame, trained on the synthesized datasets and demonstrate its effectiveness in generating React code via the $\text{pass}@k$ metric.
Our results suggest that a code VLM trained to interpret images before code generation may achieve better performance.
\end{abstract}

\section{Introduction}
The astonishing progresses of large language models (LLMs) have ushered a new era of automatic code generation~\citep{chen2021evaluating,fried2022incoder,nijkamp2022codegen,li2022competition,zheng2023codegeex,roziere2023code,li2023starcoder,luo2023wizardcoder,phungstable,guo2024deepseek,zhu2024deepseek,team2024codegemma,hui2024qwen2coder}.
Developers can now write high-quality codes simply by a couple of instruction messages, substantially increasing productivity.
The GitHub Copilot, a commercialized code editor extension built upon the code models, has been widely acknowledged by software developers and had 1.8 million paid subscribers as reported in May 2024~\citep{ghreport}.

However, natural language instructions alone are insufficient for comprehensive human-machine interaction.
This is particularly true in FE development, where certain ideas or designs are hard to articulate precisely or require cumbersome descriptions for accurate conveyance.
In real-world FE development scenarios, UI design mockups are often necessary to facilitate developers' understanding of the web design requirements.
These requirements span various dimensions, including modular development, state management, and data-driven declarations. Accurately describing such multifaceted aspects through text alone is time-intensive and prone to ambiguity, thereby limiting the practical utility of code LLMs in these contexts.
Furthermore, state-of-the-art vision-language models (VLMs) for code generation reveal substantial limitations when applied to the intricacies of FE development. 
Models like GPT-4o, though capable of generating basic components, are unsuitable for modern front-end workflows. 
Their outputs are often static, lacking modularity, reusability, and dynamic behavior—essential for scalable and interactive user interfaces—resulting in inefficient and incompatible code that deviates from established development practices.

To bridge the gap between natural language instructions and the intricacies of front-end design, we introduce a self-reflective agentic workflow that synthesizes high-fidelity training data, enabling the development of a large vision-language model for end-to-end front-end code generation.
Our key contributions are as follows:
\begin{itemize}
[noitemsep,topsep=0pt,leftmargin=9pt]
\item \textit{Self-reflective agentic workflow for data synthesis.}
A major challenge in developing a strong large-scale VLM for front-end (FE) code generation is the scarcity of high-quality image-text data. To address this, we propose an automated pipeline that extracts, renders, and annotates self-contained FE code snippets. 
As a key contribution, we introduce three novel data synthesis methods—Evolution-Based, Waterfall-Model-Based, and Additive Development synthesis—which enable the generation of large-scale, diverse, and high-fidelity training data. 
Moreover, the proposed pipeline flexibly supports data generation for both single-image and multi-image code generation, making it broadly applicable to diverse front-end development scenarios.
\item \textit{Large Vision-Language Model for Front-End Code Generation.}
We develop Flame: \textbf{F}ront-end \textbf{l}anguage \textbf{a}ssistant with \textbf{m}ultimodal \textbf{e}xpertise, a large vision language model that integrates the code generation capability of a code LLM with the visual understanding capability of a vision encoder, enabling direct translation of design concepts into executable front-end code.
We explore optimal practice for training a VLM for FE code generation through ablation studies.
Empirical studies also suggesting that model tuned to interpret the input image before code generation can potentially greatly enhance performance, analogous to how Chain-of-Thought~\citep{wei2022chain} improves large language models (LLMs).
\item \textit{Benchmarking React Code Generation.}
To the best of our knowledge, we construct Flame-React-Eval, the first benchmarking dataset designed to assess syntactic precision, functional correctness, and visual consistency in React code generation across a range of design specifications.
\item \textit{Open Source.}
To advance the field of front-end code generation, we publicly release our data synthesis pipeline, the synthesized training data, the evaluation dataset, and the Flame models\footnote{https://github.com/Flame-Code-VLM/Flame-Code-VLM}.
\end{itemize}

\section{Related Work}
\subsection{Code Models}
The progress of LLMs for code has been made at a surprisingly fast pace these years.
Codex~\citep{chen2021evaluating} showed that LLMs were produce workable solutions to real programming problems, pioneered the use of LLMs for code. 
In the development of the open-source code LLMs in recent years, the model size has been explored from 0.5 billion to 236 billion and the scale of training code data has grown to more than 3 trillion tokens, and as a result the SOTA code model can achieve more than 90\% pass@1 on the HumanEval dataset.
Besides the work on open LLMs for code, techniques for code instruction-tuning have emerged. 
The Self-Instruct~\citep{wang2022self} approach generates instruction-tuning datasets through an iterative process: prompting a powerful LLM with samples from a seed task pool, generating instructions and corresponding responses, filtering the results, and progressively expanding the task pool.
It is widely adopted in the code generation community.
For example, CodeAlpaca~\citep{codealpaca,alpaca} and CodeLlama~\citep{roziere2023code} were instruction-tuned on the datasets constructed under variants of the dea.
WizardCoder~\citep{luo2023wizardcoder} employed an evolutionary algorithm~\citep{xu2024wizardlm} to create diverse and complex
code instruction data.
OctoCoder~\citep{muennighoff2023octopack} showed that instructions dataset made from Git commits was also beneficial for code generation performance.
The advancements in LLMs for code generation have paved the way for AI software engineers. 
Leveraging the coding capabilities of LLMs, autonomous multi-agent systems such as MetaGPT~\citep{hong2024metagpt,hong2024data}, BabyAGI~\citep{babyAGI}, and ChatDev~\citep{chatdev} have demonstrated the ability to accomplish complex tasks, like writing a CLI blackjack game. 

\subsection{Large Vision-Language Models}
To fully leverage the reasoning capability of an LLM and the visual perception capability of a vision model, Flamingo~\citep{alayrac2022flamingo} was proposed connecting them to build a vision-language model. 
The design of linking a vision model to an LLM rapidly emerged as the dominant approach, with efforts on exploring various methods such as tuning strategies and techniques for perceiving high-resolution images~\citep{liu2024visual,li2023blip,zhu2023minigpt,wang2023cogvlm,bai2023qwen,chen2024internvl,lu2024deepseek,hu2024minicpm,li2024llava}. 
This architecture can be further extended to incorporate video as an additional input modality by treating it as a sequence of images
~\citep{maaz2023video,lin2023video,li2023llama,weng2024longvlm,chen2023videollm}.

\subsection{Multimodal Solution to Front-end Code Generation}
The idea of generating front-end codes via multimodal models is not new. 
The screen-to-code project~\citep{screenshot2code} was initiated to convert screenshots, mockups and Figma designs into clean, functional code by utilizing large multimodal LLMs like GPT-4o or Claude Sonnet 3.5. 
The front-end coding process is implemented through a multi-stage workflow rather than an end-to-end approach where multimodal LLMs are first employed to output the textual interpretations of input images and subsequently these textual outputs combined with instructions are utilized for code generation.
However, such generated code often lacks modular design and sophisticated state management, features critical for aligning with industry-standard development practices.
Furthermore, the Design2Code initiative~\citep{si2024design2code} advances this approach by directly converting design images into executable code through a multimodal LLM. This method enhances the integration of visual inputs and coding instructions but still primarily produces static code. 
It tends to encapsulate functionality within a single large component without supporting modular architecture or interactive elements like state management, as the limitation observed in GPT-4o. This approach falls short of the modular and interactive capabilities typically expected in contemporary FE development, highlighting the ongoing challenge of creating truly industry-ready automated coding solutions.
\cite{zhou2024bridging} proposed DeclarUI, a pipeline-based framework that generates declarative UI code from design mockups and screenshots.
DeclarUI relies on an iterative refinement process, integrating computer vision techniques, Page Transition Graphs (PTGs), and multimodal LLMs to first interpret designs and then generate structured code through multiple steps.
However, its reliance on intermediate representations and iterative optimization introduces additional complexity, potentially reducing efficiency and limiting generalization across UI frameworks.

\begin{figure*}[t]
    \centering
    \includegraphics[width=\textwidth]{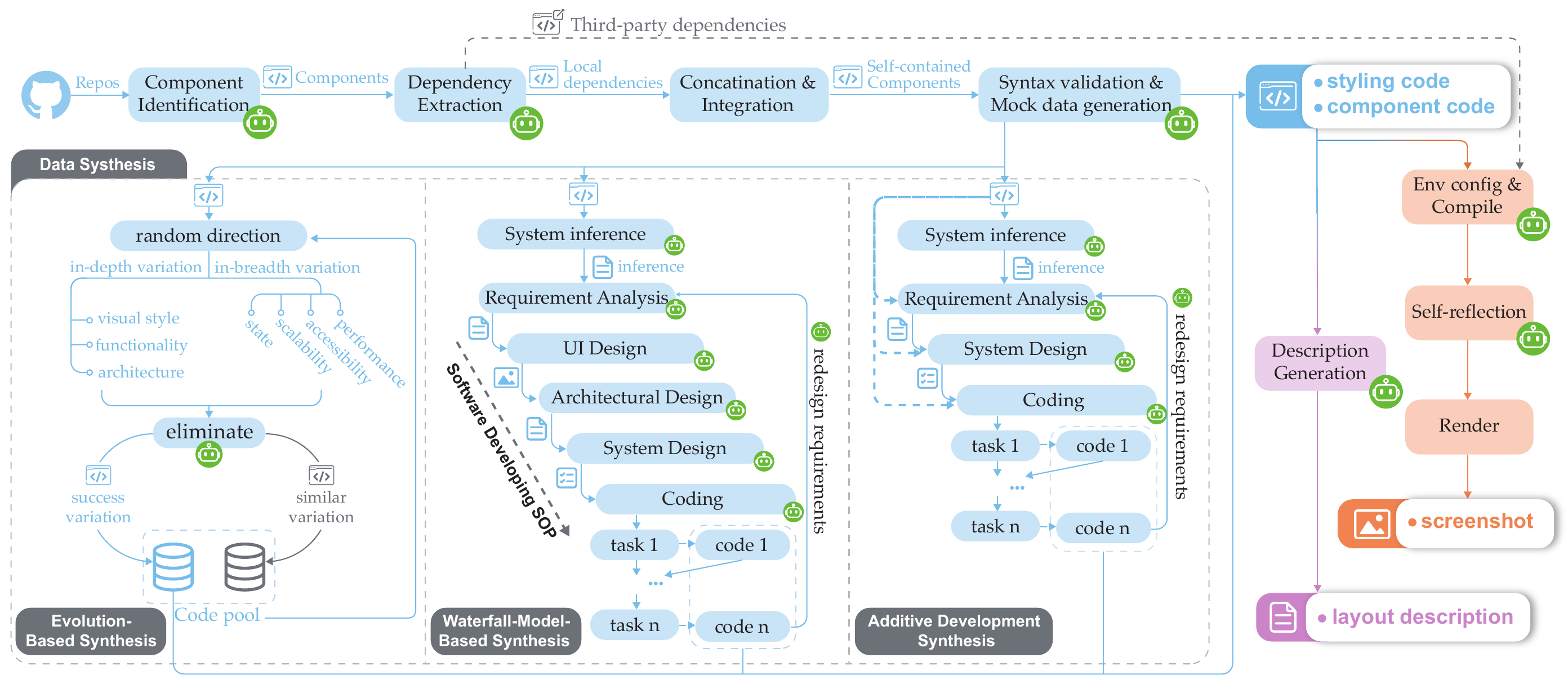}
    \caption{Overview of the data preparation pipeline. The process involves extracting self-contained code snippets from GitHub repositories and applying three code synthesis methods (blue) to enhance dataset diversity. Extracted snippets are then rendered to generate corresponding visual representations (orange), followed by the generation of structured layout descriptions for the components (purple).}
    \label{fig:data-pipeline-augmentation}
\end{figure*}

\section{Automated Multimodal Data Generation}

To ensure broad applicability across front-end development frameworks, our proposed methodology is designed to be flexible and adaptable, enabling efficient collection and generation of training data for various frameworks with minimal transition cost. 
To demonstrate its effectiveness, we apply it to React, the most widely adopted front-end framework, known for its component-based architecture that emphasizes modularity, reusability, and maintainability. 

In React, a component is a self-contained, reusable unit of UI that encapsulates both structure (JSX), behavior (JavaScript logic), and styling (CSS), facilitating a structured and scalable approach to building interactive applications. 
React also provides state management to handle dynamic data changes, one-way data flow for predictable UI updates, and a declarative rendering paradigm that simplifies UI logic by automatically reflecting state changes. 

The proposed pipeline for synthesizing a diverse and structured dataset for front-end development comprises three primary stages: Self-Contained Code Snippet Generation, Code Snippet Rendering, and Description Generation, as illustrated in Fig. \ref{fig:data-pipeline-augmentation}. 
Each synthesized data instance includes the component(s) code, styling code, layout description, and a rendered screenshot, ensuring comprehensive multimodal representation of front-end structures.
The following subsections provide further details.

\subsection{Self-Contained Code Snippet Generation}

This stage focuses on curating and preparing React components from a wide range of public repositories. 
The components span various scales, ranging from atomic elements to complex groups and entire pages. 
These components are tailored to meet the advanced requirements of training code-generating models, ensuring independence from their original development contexts and alignment with real-world application scenarios.

\subsubsection{Data Collection}

The data collection process begins with identifying React repositories using the GitHub API. 
Repositories were filtered based on predefined keywords in their descriptions, readmes, or names, while excluding boilerplate or instructional repositories using a blacklist (e.g., "learn," "tutorial," "example," "demo"). 
To ensure relevance and quality, only repositories with over 10 stars were included. 

\subsubsection{Component Identification}

Components were identified through an automated extraction process, leveraging advanced regular expressions to detect key React-specific features such as hooks, class-based components, and functional components. To enhance accuracy, an agent-based validation mechanism was employed to verify and refine the detected components. Once extracted, the components underwent parsing and preprocessing to remove irrelevant code while ensuring they were correctly paired with their respective dependencies and styles. This preprocessing step preserved the functional integrity of the components, enabling them to operate independently of their original development contexts.

\subsubsection{Code Snippet Generation}

Building upon the extracted components, this step involved transforming them into fully self-contained units suitable for multimodal LLM training. The transformation comprised three key steps:
\begin{itemize}[noitemsep,topsep=0pt,leftmargin=12pt]
\item Integration of Dependencies: Import statements were replaced with inline code, and all necessary dependencies were embedded directly within the snippet. This eliminated external references, ensuring that each component remained executable in isolation.
\item Validation and Correction: A hybrid approach combining rule-based validation and agent-assisted correction was applied to enforce syntactical and functional consistency. This step detected and resolved common issues such as formatting inconsistencies, missing references, and incorrect bindings.
\item Simulating Real-World Inputs: To enhance practical applicability, components were analyzed to identify input parameters (e.g., props and state values). An iterative, agent-driven approach was then employed to generate realistic mock data and placeholder functions, ensuring that the components could operate independently and simulate real-world use cases effectively.
\end{itemize}

\subsection{Code Data Synthesis}

While extracting self-contained code snippets from public repositories provides a valuable foundation, the quantity of collected data remains insufficient for training a large-scale vision-language model. To address this limitation, we developed a systematic data synthesis strategy incorporating three distinct synthesis methods: \textbf{Evolution-Based Synthesis}, \textbf{Waterfall-Model-Based Synthesis}, and \textbf{Additive Development Synthesis}, each designed to increase dataset volume while preserving diversity and relevance.

\subsubsection{Evolution-Based Synthesis}
This method draws inspiration from the Evol-Instruct methodology introduced in WizardLM~\citep{Xu2023WizardLMEL}. This approach employs random evolutionary logic to generate diverse variations of existing data instances. The key characteristics of this method include its adaptability to different coding styles and its ability to explore a wide spectrum of possibilities. 

Similar to Evol-Instruct, our method was also conducted by employing two evolutionary directions: in-breadth and in-depth evolution, facilitated by LLMs to ensure a broad and nuanced transformation of the initial dataset, as illustrated in Fig.\ref{fig:data-pipeline-augmentation}. 
The process began with the code snippets we collected from GitHub as the seed, subjected to iterative rounds of evolution to derive new, diversified versions encompassing a wider spectrum of potential applications.

\textbf{In-Breadth Evolution} This aspect of evolution focuses on increasing dataset diversity by systematically generating code variations that expand functionality, modify architectural structures and refine visual styles to enhance usability and adaptability. 

\textbf{In-Depth Evolution} The in-depth direction, on the other hand, deepens the technical complexity of code snippets to challenge and refine component handling, by enhancing the state management, optimizing the performance, or enforcing accessibility standards.

Applying our Evolution-based synthesis resulted in a substantial increase in the volume and diversity of the dataset.

\subsubsection{Waterfall-Model-Based Synthesis}
The Waterfall-Model-Based Synthesis employs a structured and sequential framework inspired by traditional software development methodologies. Unlike the randomness of earlier approaches, this method follows a systematic progression akin to the waterfall model, ensuring cohesion, logical structure, and alignment with real-world development practices. As illustrated in Fig.\ref{fig:data-pipeline-augmentation}, the process begins with inferring about a system or application with a given code snippet from the dataset collected from GitHub. Then, it proceeds through a series of well-defined stages:
\begin{itemize}[noitemsep,topsep=0pt,leftmargin=12pt]
\item Requirement Analysis: The broader system requirements are inferred, defining the overall functionality, user needs, and scope of the inferred system.
\item UI and Architectural Design: Based on the requirements, detailed UI layouts and architectural blueprints are developed, ensuring that the design is both functional and scalable.
\item System Design: This process refines the overall system structure by integrating the architectural and functional requirements into a cohesive design framework.
\item Coding: The design is implemented through a stepwise approach, where each step corresponds to a development task and the LLM accomplish each task by generating self-contained and functional code snippets, which contributes to the overall functionality.
\end{itemize}

This structured approach results in cohesive and logically connected code components, where each phase builds on the outcomes of the previous one. The resulting data reflect the complexities and interdependencies seen in real-world systems, enhancing the robustness and relevance of the generated code. Furthermore, the interconnected nature of the components increases their applicability across diverse scenarios.

\subsubsection{Additive Development Synthesis}
The third method prioritizes realism by leveraging human-authored code sourced from the dataset collected on GitHub. By additively building upon existing, manually crafted components, this approach ensures the generated outputs remain firmly grounded in real-world development practices while incorporating meaningful complexity and functionality, 

As illustrated in Fig.\ref{fig:data-pipeline-augmentation}, the process also begins with a human-written code snippet and follows a streamlined waterfall model. Unlike the full Waterfall-Model-Based approach, this method incorporates the given code snippet in the prompt of each step. Rather than generating systems from scratch, it develops a step-wise plan that retains the style and context of the original component. The focus is on iterative refinement and additive extension, with each cycle representing a realistic development task. The agent in each step is prompted to adapt to the coding style of the provided snippet, ensuring that the newly generated code aligns with the original style and context. For example, a button component might be incrementally enhanced with state management capabilities, accessibility improvements, or dynamic functionalities such as API integrations.

Each iteration is carefully scoped to align with practical development workflows. Tasks may include enriching components, integrating accessibility standards, or applying advanced architectural patterns. This iterative approach ensures that the final outputs are scenario-specific, reflective of human development practices, and logically cohesive. Furthermore, the incremental evolution allows for modular complexity, with changes that systematically build upon prior iterations.

\subsection{Rendering}

\subsubsection{Automatic Rendering Pipeline}
A controlled environment replicated typical React development conditions. Refined snippets were integrated into a standardized boilerplate React application, ensuring consistency and reproducibility. Automated scripts handled dynamic server configuration, dependency installation, and routing to component URLs for rendering and capturing visual outputs.

\subsubsection{Error Handling and Self-reflective Correction}

In the event of server errors, dependency installation issues, or failed component execution, a reflective agent-based strategy is deployed to analyze and resolve the issues. This method enhances the robustness of the rendering process and reduces manual intervention:

\textbf{Dependency Installation Errors}: If a dependency installation fails or takes an abnormally long time, the logs from the installation process are recorded. The agent will analyze the logs and suggest refined dependency versions or configurations. The system will attempt to re-install the dependencies, based on the agent's suggestions, with a maximum of \textit{n} retries. This approach ensures that most errors related to dependency versions or conflicts are resolved automatically.

\textbf{Component Rendering and Server Start Errors}: Errors during server startup or component rendering are similarly analyzed by the agent. Logs are examined to diagnose implementation issues, and the agent generates corrective code modifications. These changes are applied, and the process is retried iteratively, up to \textit{m} times if needed, to ensure successful execution.

This layered approach allows for self-reflection and refinement of the process, enhancing the overall efficiency and reliability of component rendering, while minimizing manual interventions and improving the accuracy of the final dataset.
\subsection{Layout Description Generation}
Visual representations of components were enriched with detailed layout descriptions.
The descriptions provided insights into design and usability by analyzing structural hierarchy, spatial alignment, and interactions among UI elements. 
Features such as alignment strategies and the relationship between interactive and static elements were documented.
These descriptions were integrated with the corresponding code snippets and visual outputs, creating a dataset enriched with contextual information that supports the training and evaluation of vision-language models.

The sizes of the datasets synthesized using the Evolution-Based, Waterfall-Model-Based, and Additive-Development methods are 164K, 175K, and 69K, respectively (see \ref{apdx:data_stat}).

\section{Modeling and Training}\label{sec:model}
In order to fully take advantage of the generation power of LLMs, the structure of the proposed model follows the mainstream design of large vision and language models composed by a vision encoder, a vision-to-text token connector and a large language model~\citep{liu2024visual,li2024llava,chen2024internvl}.
We adopted SigLIP SO400M/14@384~\citep{zhai2023sigmoid} as the vision encoder as it outperformed other candidates in our preliminary studies, aligning with the observation in existing study~\citep{tong2024cambrian}.
We choose the DeepSeek coder models~\citep{guo2024deepseek} as the code LLM bases, given their exceptional performance among code models in the open-source landscape.
The vision-to-text token connector was implemented with a 2-layer multilayer perceptron (MLP).

As a result of the model design, different training strategies have been explored and discussed in existing literature~\citep{li2024llava,tong2024cambrian,chen2024internvl,chen2024expanding}.
We adopted the three-stage training strategy.
At first, the vision-to-text connector is solely warmed up using the public dataset.
Given that increasing the amount of warm-up data improved results~\citep{tong2024cambrian} , we use both the BLIP558K and LLaVA Single-Image3.2M~\citep{li2024llava} for this stage.
Then the vision encoder bundled with the connector are trained using the synthesized dataset where the model learns to describe the layout of an input image.
This stage corresponds to the ViT incremental Learning stage in~\citep{chen2024expanding} which has shown widely beneficial in our experiments. 
At the final stage, the whole model is instruct-tuned to generate code using input images.
In the last two stages, we trained the model solely on the syntheized dataset and adopted the recent proposed AnyRes method~\citep{li2024llava} to enable the model handling high-resolution images.
All the training was executed on 24$\times$H800 GPUs, where the first stage went through the data for one epoch and the following stages trained the models with a learning rate of 2e-5 and a batch size of 1536 for two epochs. 


\section{Experiments}
\subsection{Benchmark Data}
To evaluate Flame’s ability to translate visual designs into executable React code, we constructed Flame-React-Eval, a benchmark dataset of 80 manually curated test cases, making it the first publicly available benchmark for assessing vision-language models in modern front-end development with declarative frameworks like React. Unlike DeclarUI, which provides a dataset consisting solely of UI design mockups, Flame-React-Eval offers a more comprehensive evaluation framework by incorporating rendered screenshots, structured layout descriptions, and fully functional React code (including styling). This enriched dataset enables a more rigorous assessment of multimodal models in UI-to-code translation.
The dataset was created independently of the training data to ensure an unbiased evaluation. Each test case was manually designed rather than sourced from public datasets, allowing for controlled quality and diversity. The test cases capture essential aspects of front-end development, including component hierarchies, layout structures, dynamic behaviors, etc., ensuring alignment between design mockups and their expected implementations.

\subsection{$\text{pass}@k$ for Front-End Code Evaluation}
The $\text{pass}@k$ metric is commonly used in code generation evaluations as it focuses on functional correctness rather than mere code similarity~\citep{chen2021evaluating} where an implementation is counted as a correct solution when it passes the unit test.
However, in front-end development, constructing fair unit tests is challenging, since the front-end correctness is inherently tied to structure, layout, and visual fidelity, unlike algorithmic tasks where outputs are independent of implementation. Front-end unit tests enforce strict assertions on component structures and rendering behavior, restricting the diverse yet valid implementations that can achieve the same UI. They also fail to capture design accuracy, responsiveness, and styling variations, making them ineffective for assigning binary correctness in pass@k. 

In this work, a solution to an FE problem is considered correct if it meets the following criteria: 1) it successfully compiles; 2) it renders a non-error image; and 3) the rendered image from the generated code matches the reference image.
We use cosine similarity between the visual embeddings of the rendered and reference images to assess the matching result.
The embeddings are computed using the DINOv2-base model~\citep{oquab2023dinov2} and images are considered almost identical when the cosine similarity of their embeddings exceeds 0.9, a threshold that has been tested to align better with human visual perception of similarity compared to other methods.
For reference, the similarities of different image pairs are shown in appendix \ref{apdx:sim-show} for reference.
In our experiments, we generate $50$ solutions per test case and report the $\text{pass}@k$ scores for $k=1,3,5$, with the threshold of image similarity set at $0.9$.

\subsection{Image to React code} 
We evaluate the performance of Flame models in generating React code from an instruction and an image. 
During the final instruction-tuning stage, the training data is divided into two parts: one for generating code directly from the image and instruction, and the other for generating an image layout description (i.e., interpretation of the image) before producing the corresponding code.
We also compare Flame models against SOTA models, including GPT-4o, Gemini 1.5-Flash, InternVL2.5 78B~\citep{chen2024internvl}, LLaVA-Qwen2-72B-OV, and LLaVA-Qwen2-7B-OV~\citep{li2024llava}. 
In all experiments, we set the temperature to 0.1 and apply nucleus sampling with top\_p = 0.95.
Results are shown in Table \ref{tab:performance}, with testing prompts provided in the appendix \ref{apdix:train-prompt}.

Flame models trained on Waterfall-Model-Based and Additive-Development synthetic data achieve superior $\text{pass}@k$ performance compared to that trained on evolution-based synthetic data. 
This is likely because structured data synthesis aligns better with real-world coding patterns, providing more coherent and contextually meaningful training data. 
The deterministic nature of the Waterfall-Model-Based and Additive-Development methods ensures logical progression in code generation, reinforcing consistent syntax, modularity, and functional correctness. 
In contrast, the random evolving in Evolution-based synthesis may introduce noise and unnatural variations that hinder learning robust coding patterns.
Compared to the Waterfall-Model-Based method, the Additive Development method produces significantly fewer data points to train a model achieving similar performance. 
However, it heavily relies on initial code snippets within its iterative generation process. 
In contrast, the Waterfall-Model-Based method has minimal dependence on seed data but produces a much larger dataset. 
This highlights a key tradeoff in data synthesis: the Additive-Development method generates data more selectively but with a more complex process, while the Waterfall-Model-Based approach favors ease of generation and naturally results in a much larger dataset.
Additionally, both open-source and proprietary SOTA models fail to match Flame's performance, highlighting their limitations in generating React front-end code.
\begin{table}[htbp]
\centering
\small
\renewcommand{\arraystretch}{1.5}
\begin{tabular}{l c c c}
\toprule
\textbf{Model} 
& \textbf{pass@1} 
& \textbf{pass@3} 
& \textbf{pass@5} \\
\midrule
\textbf{Flame-Evo-7B} & 43.8\%  & 62.9\%  & 69.1\% \\
\textbf{Flame-Waterfall-7B} & \textbf{52.6\%}  & 65.5\%  & 70.3\% \\
\textbf{Flame-Additive-7B} & 51.6\%  & \textbf{67.0\%}  & \textbf{71.9\%} \\
\midrule
\textbf{LLaVA-Qwen2-72B-OV} & 7.3\%  & 10.5\%  & 11.9\% \\
\textbf{LLaVA-Qwen2-7B-OV} & 3.6\%  & 6.5\%  & 8.0\% \\
\textbf{InternVL2.5 78B} & 3.8\%  & 5.6\%  & 6.5\% \\
\textbf{GPT-4o-2024-08-06} & 4.9\% & 6.7\% & 7.6\% \\
\textbf{Gemini 1.5 Flash} & 11.1\% & 13.7\% & 14.5\% \\
\bottomrule
\end{tabular}
\caption{Performance comparison of different models for react code generation.}
\label{tab:performance}
\end{table}
\subsection{Ablation Studies} 
We conduct ablation studies to investigate the impact of different training strategies and data mixing recipes, with results shown in Table \ref{tab:ablation}.
The 3-Stage strategy, as described in Section \ref{sec:model}, includes a middle stage where the VLM undergoes instruction-tuning after connector warm-up, whereas the 2-Stage strategy skips this step.
The "recipe" column indicates the data mixing strategies used during the final instruction-tuning stage:
"C" refers to direct code generation from the image and instruction ("Code Only"); "IC" requires half of the tasks to generate an image layout description (i.e., interpretation) before producing the corresponding FE code ("Interpretation Before Coding" + "Code Only").
We also evaluate the performance of the LLaVA-Qwen2-7B-OV model fine-tuned (via supervised fine-tuning, SFT) on the synthesized datasets.
The results show that, when training Flame models from scratch, the 3-Stage strategy outperforms the 2-Stage strategy, suggesting that the middle stage enhances the model's ability to interpret images, thereby improving performance.
Additionally, the "Interpretation Before Coding" + "Code Only" recipe enhances the code generation performance of Flame models, further indicating that learning to interpret images strengthens the capabilities of vision-language models (VLMs).
Directly fine-tuning an off-the-shelf model like LLaVA-Qwen2-7B-OV yields performance comparable to that of Flame models, though Flame, designed for a specific image-to-code task, requires significantly less training data and effort--only 3.3 million data points--compared to approximately 6 million for LLaVA-Qwen2-7B-OV.
We also note that no significant difference exists between the two data mixing strategies for off-the-shelf models, possibly because these models are already well visual-instruction-tuned and less sensitive to new instructions.
\begin{table}[htbp]
\centering
\small
\setlength{\tabcolsep}{7pt} 
\renewcommand{\arraystretch}{1.1} 
\begin{tabular}{c llc ccc}
\toprule
\textbf{Data} & \textbf{Model} & \textbf{Mode} & \textbf{Recipe} & \textbf{pass@1} & \textbf{pass@3} & \textbf{pass@5} \\
\midrule
\multirow{7}{*}{{\textbf{Evolution}}} 
 & \multirow{4}{*}{\textbf{Flame (7B)}}
   & \multirow{2}{*}{3-Stage} & IC & 43.8\% & 62.9\% & 69.1\% \\
 &  &  & C & 42.0\% & 61.3\% & 68.6\% \\
 \cmidrule{3-7}
 &  & \multirow{2}{*}{2-Stage} & IC & 43.2\% & 59.5\% & 66.7\% \\
 &  &  & C & 38.6\% & 52.9\% & 58.4\% \\
 \cmidrule{2-7}
 & \multirow{2}{*}{\shortstack{\textbf{LLaVA-Qwen2-7B-OV} }} 
   & \multirow{2}{*}{SFT} & IC & 43.9\% & 64.3\% & 72.3\% \\
 &  &  & C & 41.2\% & 59.4\% & 67.1\% \\
\midrule
\multirow{7}{*}{{\textbf{Waterfall}}} 
 & \multirow{4}{*}{\textbf{Flame (7B)}} 
   & \multirow{2}{*}{3-Stage} & IC & 52.6\% & 65.5\% & 70.3\% \\
 &  &  & C & 47.5\% & 59.9\% & 64.2\% \\
 \cmidrule{3-7}
 &  & \multirow{2}{*}{2-Stage} & IC & 49.0\% & 63.5\% & 68.7\% \\
 &  &  & C & 47.7\% & 59.9\% & 64.2\% \\
 \cmidrule{2-7}
 & \multirow{2}{*}{\shortstack{\textbf{LLaVA-Qwen2-7B-OV}}} 
   & \multirow{2}{*}{SFT} & IC & 47.4\% & 61.0\% & 66.2\% \\
 &  &  & C & 50.2\% & 65.1\% & 70.4\% \\
\midrule
\multirow{7}{*}{{\textbf{Additive}}} 
 & \multirow{4}{*}{\textbf{Flame (7B)}}
   & \multirow{2}{*}{3-Stage} & IC & 51.6\% & 67.0\% & 71.9\% \\
 &  &  & C & 44.2\% & 58.8\% & 64.2\% \\
 \cmidrule{3-7}
 &  & \multirow{2}{*}{2-Stage} & IC & 44.7\% & 61.0\% & 66.6\% \\
 &  &  & C & 42.8\% & 57.9\% & 63.4\% \\
 \cmidrule{2-7}
 & \multirow{2}{*}{\shortstack{\textbf{LLaVA-Qwen2-7B-OV}}} 
   & \multirow{2}{*}{SFT} & IC & 46.2\% & 63.5\% & 69.8\% \\
 &  &  & C & 49.5\% & 66.0\% & 71.7\% \\
\bottomrule
\end{tabular}
\caption{Results of ablation study on training strategies and data mixing recipes.}
\label{tab:ablation}
\end{table}
\subsection{Interpreting Before Coding} 
Recognizing the potential benefits of incorporating image interpretation into the training data, we conducted experiments to explore its impact. 
We trained Flame specifically on the task of generating the layout description of the input image before producing the corresponding code, as indicated by the "Itp" suffix in Table \ref{tab:VLMCoT}. 
During inference, models were prompted to interpret the image first before generating code. 

Results show that models tuned with the interpretation-first approach on the Evolution-Based and Waterfall-Model-Based datasets perform comparably to the original Flame model in terms of $\text{pass}@1$, but achieve significantly higher scores when $k=3$ or $5$ (Table \ref{tab:performance}).
This suggests that generating code with image interpretation encourages a more structured, step-by-step reasoning process, which in turn helps the model explore the solution space more effectively.
Moreover, interpreting before coding leads to the most significant performance improvement for the model trained on the Evolution-Based synthetic dataset, which previously exhibited the weakest performance. 
This suggests that interpretation-first can effectively mitigate the limitations of the original dataset, compensating for gaps in training data quality. 
Notably, this improvement underscores the potential of structured intermediate representations in enhancing front-end code generation from images, aligning with findings from existing literature~\citep{shao2024visual}. 
For Additive-Dev data, however, interpretation-first does not perform well.
A possible explanation is that the Additive-Dev dataset is significantly smaller than the other two, potentially lacking sufficient examples for both the middle training and final tuning stages, leading to ineffective learning of image interpretation.
\begin{table}[b!]
\centering
\footnotesize
\begin{tabular}{l c c c}
\toprule
\textbf{Model}  
& \textbf{pass@1} 
& \textbf{pass@3} 
& \textbf{pass@5} \\
\midrule
\textbf{Flame-Evo-Itp (7B)} & 44.6\%  & 68.1\%  & 75.8\% \\
\midrule
\textbf{Flame-Waterfall-Itp (7B)} & 50.2\%  & 67.7\%  & 74.2\% \\ 
\midrule
\textbf{Flame-Additive-Itp (7B)} & 43.5\%  & 63.9\%  & 70.2\% \\ 
\bottomrule
\end{tabular}
\caption{Performance Comparison of Model Versions with Different Training Recipes}
\label{tab:VLMCoT}
\end{table}

\subsection{Code Generation with Multi-Image Input}
In front-end development, beyond image-to-code translation, a common practice involves the iterative refinement of implementations in response to evolving design and functional requirements. 
For instance, UI modifications are typically introduced incrementally, with a focus on preserving structural and functional integrity.
To support such iterative process, we propose a model that generates code for updated designs using inputs from both the previous version’s design mockup and implementation as well as the updated design mockup.
The proposed waterfall-model-based and additive development synthesis approach naturally support creating training data that mimics real-world iterative processes. 
They produce versioned design-code pairs through structured and incremental modifications.
Consequently, our dataset comes pre-populated with aligned samples where updated designs match refined implementations, simplifying synthesizing training data for multi-image input scenarios. 
This structured nature of the proposed approach ensures data consistency across iterations, minimizes noise, and helps the model learn realistic update patterns, enhancing its applicability to practical front-end development workflows.

We have constructed a dataset of 603K entries for multi-image input scenarios, derived from the waterfall-model-based dataset.
Additionally, we modified the benchmark data to evaluate multi-image task.
The Flame model trained on this dataset achieves $\text{pass}@k$ of 52.2\%, 61.8\% and 64.8\% for $k=1,3,5$, showing that the proposed synthesis pipeline is well-suited for the multi-image problem.
Due to page limitations, an illustration of the rendered results is provided in the appendix \ref{apdx:multi-image}.

\section{Limitations}
Despite the promising results, there remains significant room for improvement.
Firstly, while our proposed data synthesis framework ensures consistency between code implementations and rendered images, it lacks a rigorous verification procedure for layout descriptions using both the code snippet and rendered image. 
Strengthening this verification process is crucial for improving data quality and reliability.
Secondly, our current approach is limited to single-turn interactions, which constrains its practical use in real-world scenarios. However, our method demonstrates potential for constructing datasets that could support the development of multi-turn generation capabilities, an essential direction for future work.
Lastly, the scope of data synthesis is currently restricted to the React framework. Expanding its capabilities to other front-end frameworks, such as Vue and Angular, would significantly enhance its versatility and broader applicability within front-end development.

\section{Conclusion}
In this paper, we propose an automated data synthesis pipeline for building vision-language models in front-end code generation. Our pipeline, incorporating Evolution-Based, Waterfall-Model-Based, and Additive Development synthesis methods, produces diverse, high-quality datasets and is highly adaptable to various front-end development scenarios.
We further introduce Flame and find that incorporating an intermediate training stage to refine the vision encoder and connector significantly enhances overall performance. Additionally, we observe that interpreting before coding holds great potential for improving the front-end coding capabilities of VLMs.
\bibliographystyle{unsrtnat}







\appendix
\newpage
\onecolumn
\section{Appendix}
\subsection{Data statistics of the synthesized datasets }\label{apdx:data_stat}
\begin{table}[htbp]
    \centering
    \begin{tabular}{l l}
        \toprule
        Dataset & Size \\
        \midrule
        Evolution-Based & 164,860 \\
        Waterfall-Model-Based & 175,485 \\
        Additive-Dev & 69,458 \\
        \bottomrule
    \end{tabular}
    \caption{Data statistics of datasets constructed by the methods of the Evolution-Based, Wasterfall-Model-Based and Additive Development Synthesis.}
    \label{tab:data_size}
\end{table}
\subsection{An Example Illustrating the Limitation of GPT-4o}
\begin{figure}[htbp]
    \centering
    \subfloat[A simple screenshot of a webpage.]{
        \includegraphics[width=0.31\textwidth]{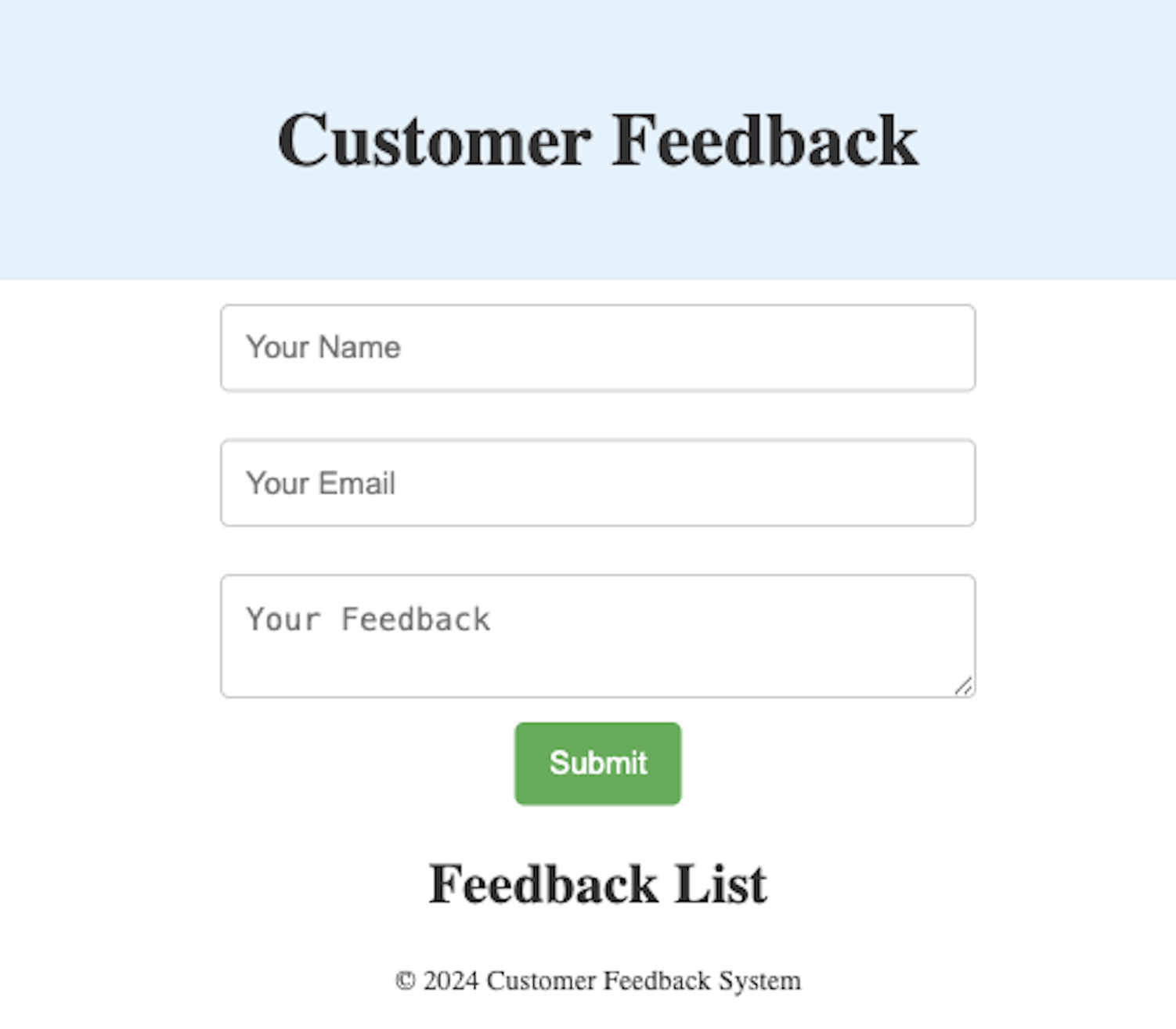}
        \label{fig:example-page}
    }
    \hfill
    \subfloat[Image rendered from the code generated by GPT-4o.]{
        \includegraphics[width=0.31\textwidth]{example_1}
        \label{fig:gpt-image}
    }
    \hfill
    \subfloat[Image rendered from the code generated by Flame.]{
        \includegraphics[width=0.31\textwidth]{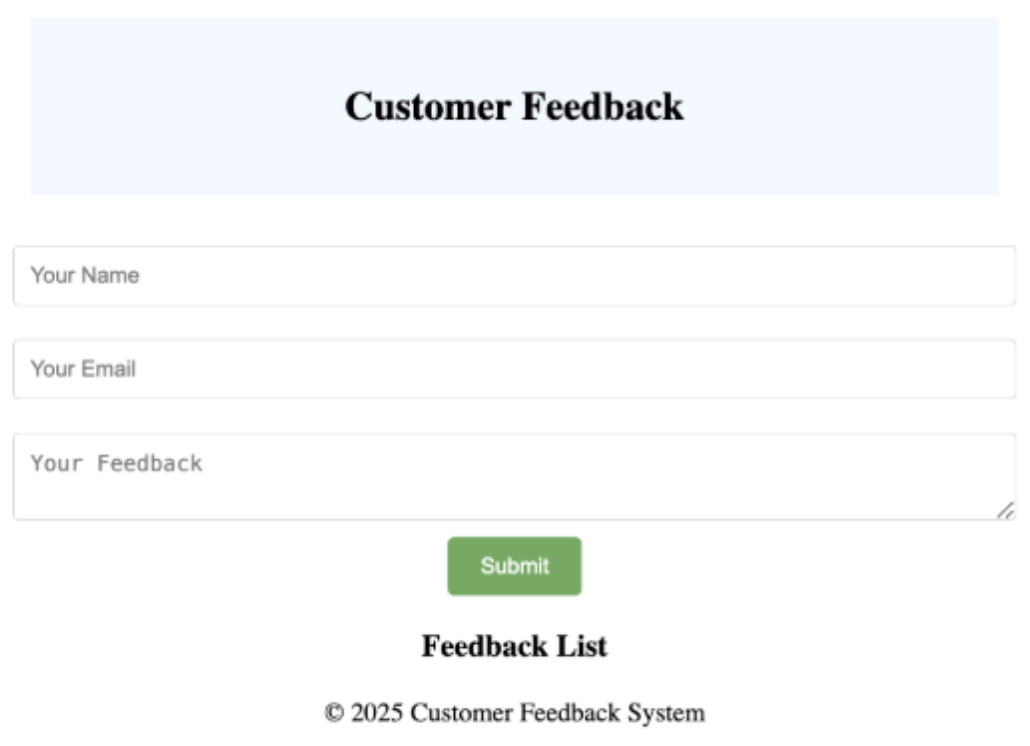}
        \label{fig:flame-image-1}
    }
    
\end{figure}

Given an example page as shown in Fig.\ref{fig:example-page}. We are expecting the React implementation to be like the following:

The above implementation effectively illustrates several defining characteristics of the React framework, emphasizing its component-based architecture, declarative UI paradigm, and state-driven rendering mechanism. These features collectively underscore React's efficacy in building modern, scalable, and maintainable user interfaces.


\begin{figure}[htbp]
    \centering
    \subfloat[The code generated by Flame for Fig\ref{fig:example-page}]{
        \includegraphics[width=0.48\textwidth]{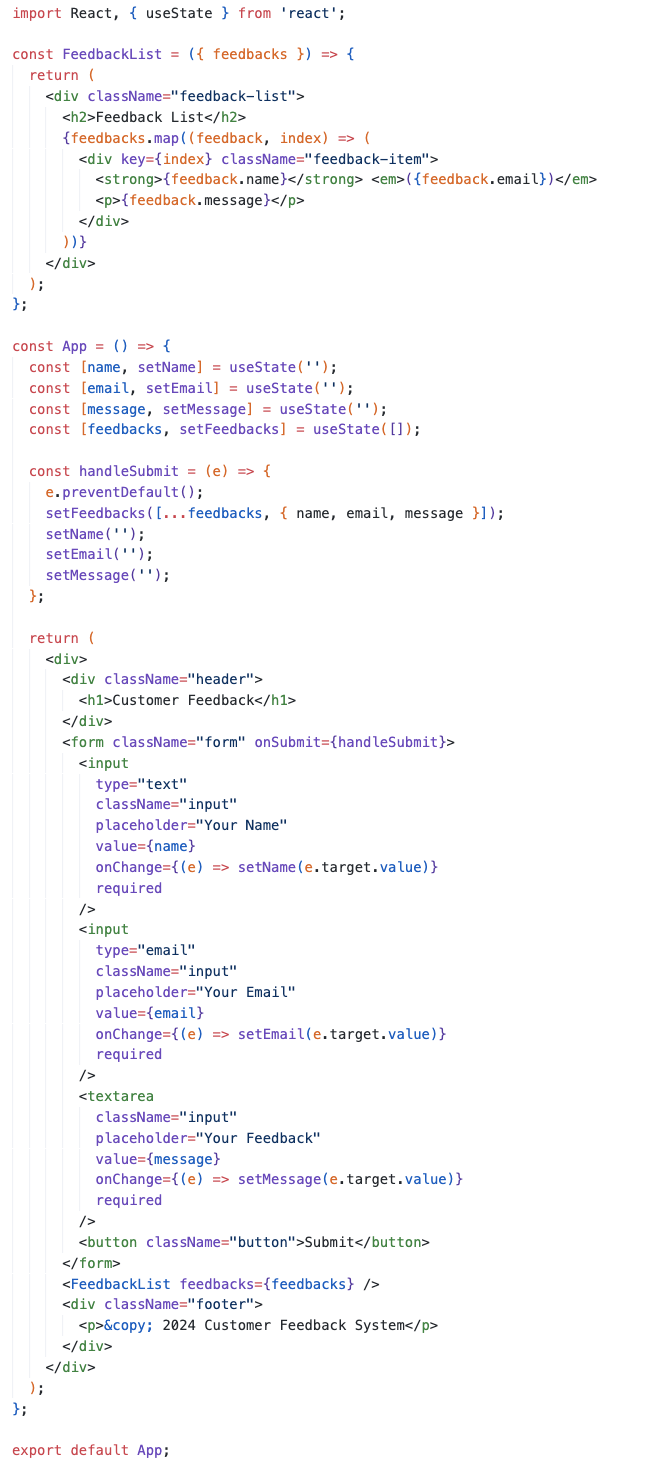}
        \label{fig:flame-result-1}
    }
    \hfill
    \subfloat[Code generated by GPT-4o for Fig.~\ref{fig:example-page}.]{
        \includegraphics[width=0.48\textwidth]{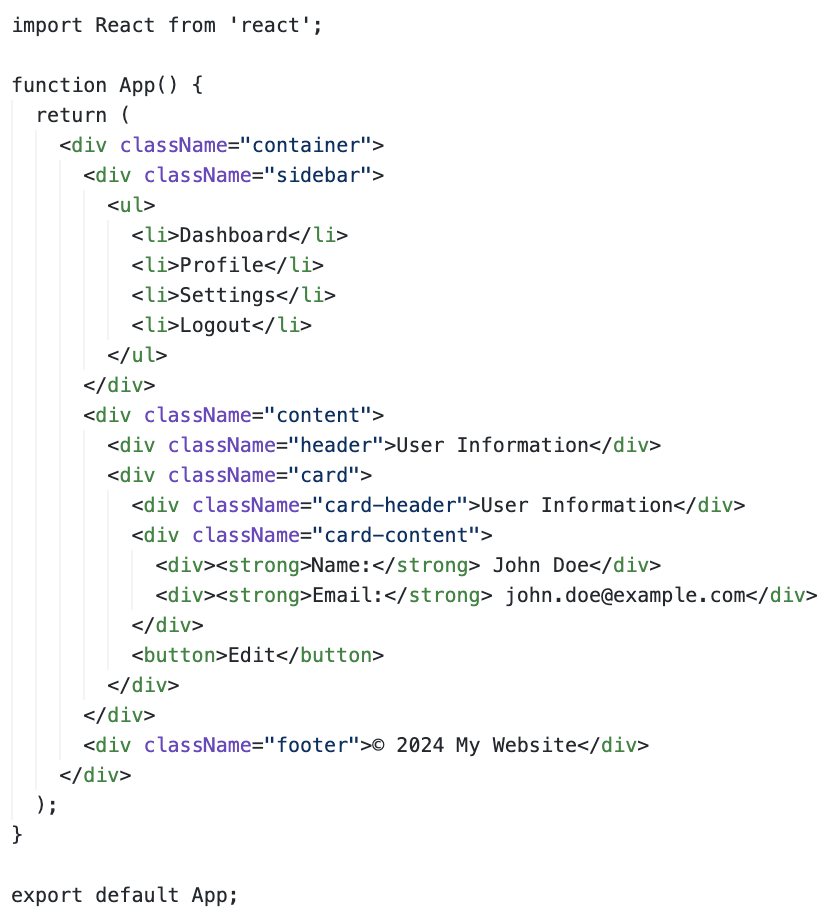}
        \label{fig:gpt-code}
    }
    \caption{Comparison of reference image and GPT-4o generated result.}
    \label{fig:comparison-gpt-code}
\end{figure}

React's component-based architecture enables the creation of modular and reusable UI elements. In this example, the FeedbackList component renders feedback items received via the feedbacks prop, allowing reuse across different parts of the application or separate projects. The App component, which manages form state and submission logic, illustrates how components can be composed to build complex interfaces while maintaining a clear separation of concerns.

A key strength of React is its declarative approach to UI development. Instead of imperatively manipulating the DOM, developers define UI states with JSX, and React efficiently updates the UI as data changes. For instance, FeedbackList dynamically renders feedback items based on the feedbacks array, ensuring automatic UI updates and improving code clarity, maintainability, and error prevention.

React’s state-driven rendering ensures synchronization between UI and application state. The App component utilizes useState to manage form inputs and feedback lists. When the form is submitted, state updates trigger a re-render, eliminating manual DOM manipulation and streamlining development. This approach enables dynamic and responsive interfaces aligned with React’s efficient reactivity model.

However, directly generating React code from screenshots using state-of-the-art (SOTA) vision-language models like ChatGPT-4o introduces significant shortcomings that conflict with React’s best practices.

Firstly, the generated code is monolithic, with the entire UI constructed within a single App component, making it difficult to maintain or extend. Essential UI elements—sidebar, content, and footer—are hardcoded together rather than structured as separate, reusable components (e.g., Sidebar, Header, Card, Footer). A modular approach would improve readability, flexibility, and scalability.

Secondly, the generated implementation lacks state-driven interactivity. User details, such as name and email, are hardcoded in JSX, preventing real-time updates. A better approach would use state variables and allow interactions like editing user data via a modal. This would align with React’s state-driven paradigm, making the UI dynamic and responsive.

Additionally, the code resembles static HTML rather than a dynamic React application. There is no clear separation between data and presentation, reducing maintainability. A declarative approach would pass user data as props to a Card component, ensuring UI elements automatically update based on state changes.

Furthermore, without proper state management and reusable components, the implementation is difficult to extend or modify. Adding new features, such as profile editing or a dynamic sidebar menu, would require significant refactoring. A well-structured React application should follow a clear component hierarchy, use useState or context, and, for complex applications, incorporate a state management library.

\subsection{An Example of Self-Contained Code Snippet}
\begin{figure}[htbp]
    \centering    \includegraphics[width=0.8\columnwidth]{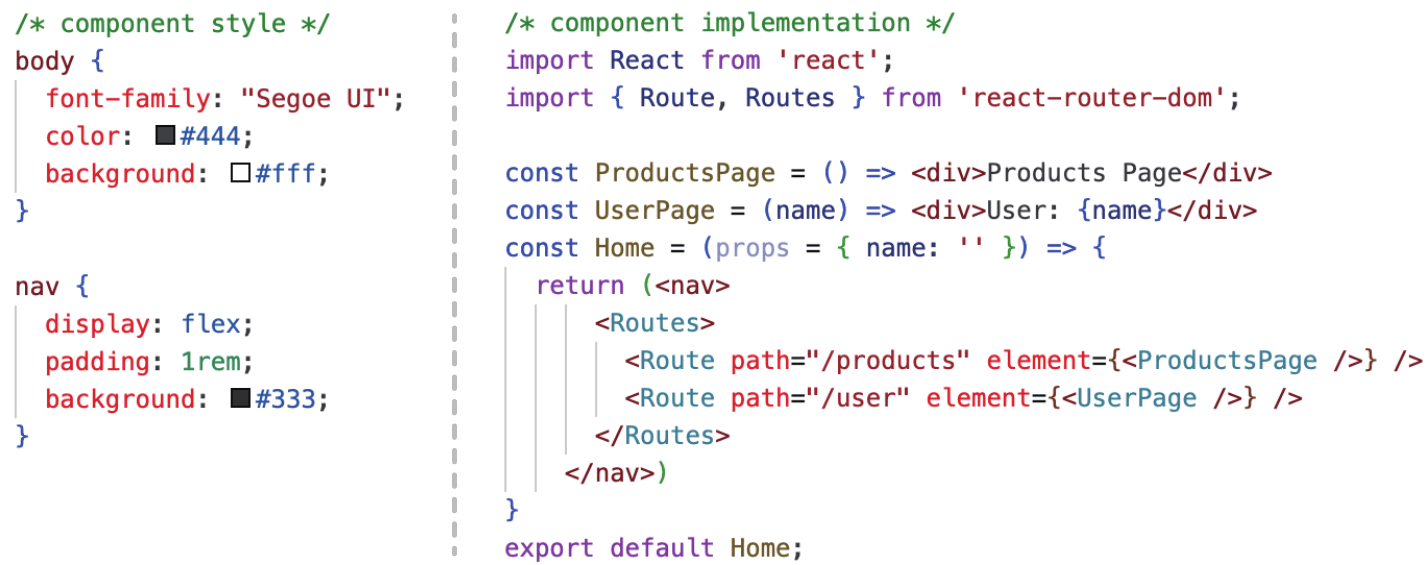}
    \caption{Self-contained code snippet containing both the component style (left) and implementation code (right).}
    \label{fig:code-snippet-eg}
\end{figure}

\subsection{An Example of Synthesized Image-Text Instance}
We show a data instance created from the proposed data synthesis pipeline.
Each synthesized data instance includes the component(s) code, styling code, layout description, and a rendered screenshot, ensuring comprehensive multimodal representation of front-end structures.
Each instance includes includes the component(s) code (Component Code), styling code (Style Code), layout description (Layout description), and a rendered screenshot.

\begin{figure}[htbp]
    \centering    \includegraphics[width=0.8\columnwidth]{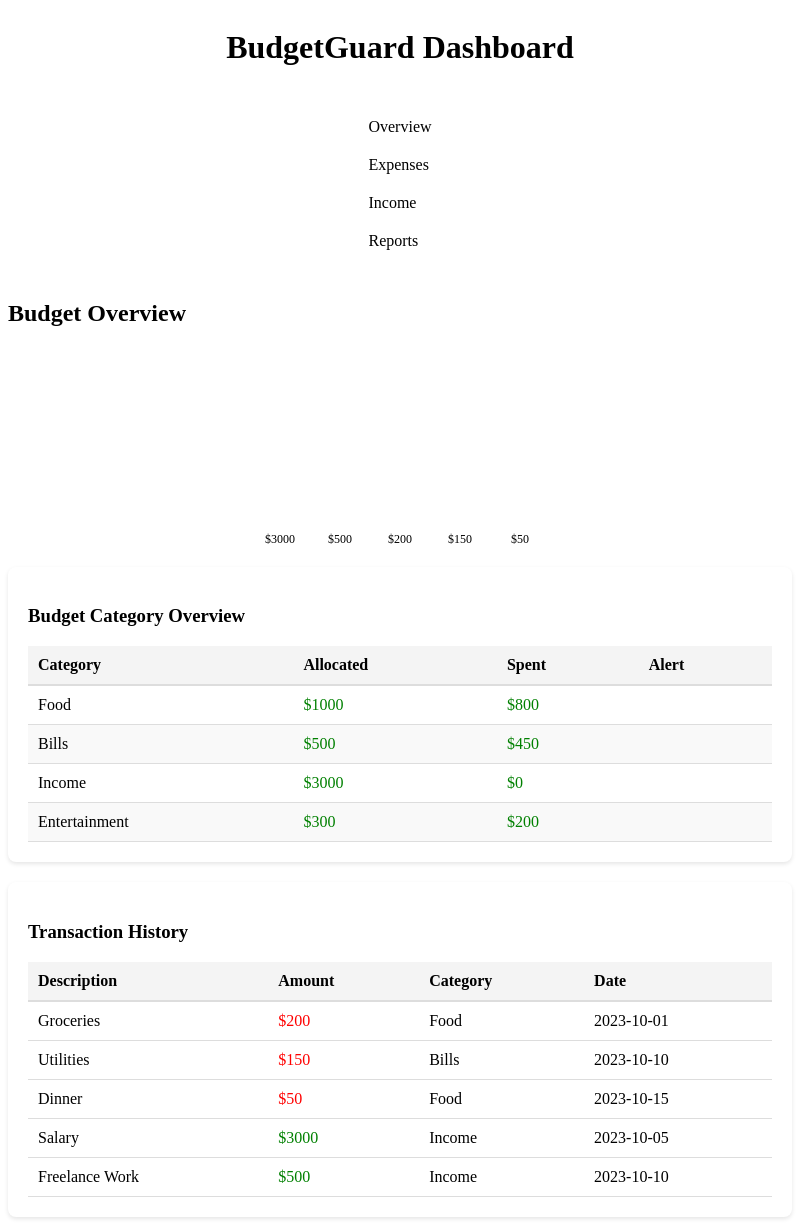}
    \caption{Screenshot of the example of synthesized image-text instance}
    \label{fig:data-sample}
\end{figure}

\begin{lstlisting} 
[Layout description]: 
The layout consists of a banner that informs users about the use of cookies and provides options to accept, reject, or customize preferences. The banner appears if no prior consent has been recorded and includes a message about cookie usage, a link to a privacy policy, and toggle options for different types of cookies. Users can interact with the banner to save their preferences, which are stored for future visits. Below the banner, a legal information page is structured to display recent updates, a search feature for easy navigation, and a list of collapsible sections containing detailed legal content. Each section can be expanded or collapsed for better readability, and one section includes an agreement checkbox for user acknowledgment. A feedback form is provided for users to share their thoughts, and a print option allows the page to be printed. The page also shows a version history for transparency. The design adapts to a dark mode setting, ensuring visual consistency across all elements. The layout is interactive, user-friendly, and organized to provide clear access to legal information and user controls.

[Task description]:
The system shall display a cookie consent banner to inform users about cookie usage and provide options to accept, reject, or customize preferences, with choices stored in local storage for persistence. A legal information page shall be included, featuring collapsible sections for terms of service, privacy policy, and accessibility, along with a search bar for easy navigation and a notification banner for recent updates. Users shall be able to provide feedback through a dedicated form and print the legal content for offline use. The interface shall adapt to a dark mode setting for accessibility and user preference, ensuring a consistent and visually cohesive experience. All interactive elements, such as buttons, checkboxes, and collapsible sections, shall be functional and user-friendly to enhance usability.

[Style Code]:
// CSS
.cookie-banner {
  position: fixed;
  bottom: 0;
  left: 0;
  right: 0;
  background-color: #ffffff;
  padding: 1rem;
  border-top: 1px solid #e0e0e0;
  display: flex;
  justify-content: space-between;
  align-items: center;
  z-index: 1000;
}

.cookie-banner.dark {
  background-color: #2d2d2d;
  border-color: #444444;
  color: #ffffff;
}

.cookie-banner button {
  padding: 0.5rem 1rem;
  border: none;
  border-radius: 0.25rem;
  cursor: pointer;
  margin-left: 0.5rem;
}

.cookie-banner button.accept {
  background-color: #3b82f6;
  color: #ffffff;
}

.cookie-banner button.reject {
  background-color: #ef4444;
  color: #ffffff;
}

.cookie-banner button.manage {
  background-color: #e0e0e0;
  color: #000000;
}

.cookie-banner button.dark {
  background-color: #60a5fa;
}

.cookie-banner button.reject.dark {
  background-color: #dc2626;
}

.cookie-banner button.manage.dark {
  background-color: #444444;
  color: #ffffff;
}

[Component Code]: 
import React, { useState, useEffect } from 'react';

const CookieConsentBanner = ({ isDarkMode }) => {
  const [isVisible, setIsVisible] = useState(false);
  const [preferences, setPreferences] = useState({
    analytics: false,
    marketing: false,
  });

  useEffect(() => {
    const consent = localStorage.getItem('cookieConsent');
    if (!consent) {
      setIsVisible(true);
    }
  }, []);

  const handleAccept = () => {
    localStorage.setItem('cookieConsent', 'accepted');
    setIsVisible(false);
  };

  const handleReject = () => {
    localStorage.setItem('cookieConsent', 'rejected');
    setIsVisible(false);
  };

  const handleManage = () => {
    console.log('Manage preferences:', preferences);
  };

  const handlePreferenceChange = (type) => {
    setPreferences((prev) => ({
      ...prev,
      [type]: !prev[type],
    }));
  };

  if (!isVisible) return null;

  return (
    <div className={`cookie-banner ${isDarkMode ? 'dark' : ''}`}>
      <div>
        <p>
          We use cookies to enhance your experience. By continuing to visit this site, you agree to our use of cookies. 
          <a href="#privacy" style={{ marginLeft: '0.5rem', color: isDarkMode ? '#60a5fa' : '#3b82f6' }}>Privacy Policy</a>
        </p>
        <div style={{ marginTop: '0.5rem' }}>
          <label>
            <input
              type="checkbox"
              checked={preferences.analytics}
              onChange={() => handlePreferenceChange('analytics')}
            />
            Analytics
          </label>
          <label style={{ marginLeft: '1rem' }}>
            <input
              type="checkbox"
              checked={preferences.marketing}
              onChange={() => handlePreferenceChange('marketing')}
            />
            Marketing
          </label>
        </div>
      </div>
      <div>
        <button onClick={handleAccept} className={`accept ${isDarkMode ? 'dark' : ''}`}>Accept</button>
        <button onClick={handleReject} className={`reject ${isDarkMode ? 'dark' : ''}`}>Reject</button>
        <button onClick={handleManage} className={`manage ${isDarkMode ? 'dark' : ''}`}>Manage Preferences</button>
      </div>
    </div>
  );
};

const LegalPage = ({ isDarkMode }) => {
  const [searchTerm, setSearchTerm] = useState('');
  const [feedback, setFeedback] = useState('');
  const [collapsedSections, setCollapsedSections] = useState({});
  const [isAgreed, setIsAgreed] = useState(false);

  const handleSearch = (e) => {
    setSearchTerm(e.target.value);
  };

  const handleFeedbackSubmit = (e) => {
    e.preventDefault();
    console.log('Feedback submitted:', feedback);
  };

  const toggleSection = (sectionId) => {
    setCollapsedSections((prev) => ({
      ...prev,
      [sectionId]: !prev[sectionId],
    }));
  };

  const handlePrint = () => {
    window.print();
  };

  const sections = [
    { id: 'terms', title: 'Terms of Service', content: 'This is the content for the Terms of Service.' },
    { id: 'privacy', title: 'Privacy Policy', content: 'This is the content for the Privacy Policy.' },
    { id: 'accessibility', title: 'Accessibility', content: 'This is the content for Accessibility.' },
  ];

  return (
    <div className={`legal-page ${isDarkMode ? 'dark' : ''}`}>
      <div className={`notification-banner ${isDarkMode ? 'dark' : ''}`}>
        Recent updates: Updated Privacy Policy (Last Updated: 2023-10-01)
      </div>
      <h1>Legal Pages</h1>
      <div className="search-bar">
        <input
          type="text"
          placeholder="Search legal pages..."
          value={searchTerm}
          onChange={handleSearch}
        />
      </div>
      <div className="table-of-contents">
        <ul>
          {sections.map((section) => (
            <li key={section.id}>
              <a href={`#${section.id}`}>{section.title}</a>
            </li>
          ))}
        </ul>
      </div>
      {sections.map((section) => (
        <div key={section.id} className="collapsible-section">
          <div
            className={`collapsible-section-header ${isDarkMode ? 'dark' : ''}`}
            onClick={() => toggleSection(section.id)}
          >
            <h2 id={section.id}>{section.title}</h2>
            <span>{collapsedSections[section.id] ? '\u25bc' : '\u25b2'}</span>
          </div>
          {!collapsedSections[section.id] && (
            <div className="collapsible-section-content">
              <p>{section.content}</p>
              {section.id === 'terms' && (
                <div className="terms-checkbox">
                  <input
                    type="checkbox"
                    id="agree"
                    checked={isAgreed}
                    onChange={(e) => setIsAgreed(e.target.checked)}
                  />
                  <label htmlFor="agree">I agree to the Terms of Service</label>
                </div>
              )}
            </div>
          )}
        </div>
      ))}
      <div className="feedback-form">
        <form onSubmit={handleFeedbackSubmit}>
          <textarea
            placeholder="Provide your feedback..."
            value={feedback}
            onChange={(e) => setFeedback(e.target.value)}
          />
          <button type="submit" className={isDarkMode ? 'dark' : ''}>Submit Feedback</button>
        </form>
      </div>
      <div className="print-button">
        <button onClick={handlePrint} className={isDarkMode ? 'dark' : ''}>Print Page</button>
      </div>
      <div className="version-history">
        <p>Version History:</p>
        <p>Last Updated: 2023-10-01</p>
      </div>
      <Chatbot isDarkMode={isDarkMode} />
      <CookieConsentBanner isDarkMode={isDarkMode} />
    </div>
  );
};

export default LegalPage;
\end{lstlisting}

\subsection{A Generation Example of Flame with Multi-Image Input}\label{apdx:multi-image}

An example scenario where the model takes the screenshot and implementation code of the previous version of the system as well as the design mock-up of the desired system as the input, and generates the implementation code correspond to the design mock-up based on current implementation. 

\begin{figure}[htbp]
    \centering
    \subfloat[Screenshot of previous version]{
        \includegraphics[width=0.3\textwidth]{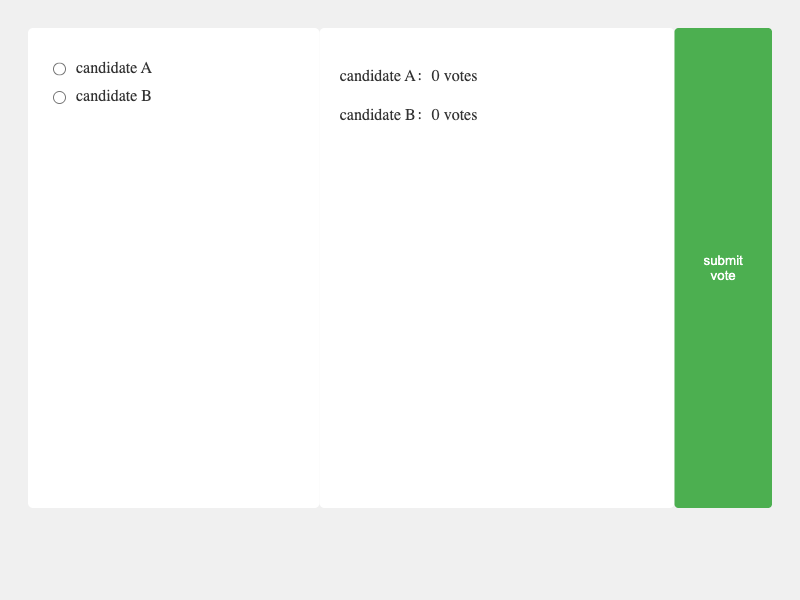}
        \label{fig:multi-img1}
    }
    \hfill
    \subfloat[Design mock-up of the desired system]{
        \includegraphics[width=0.3\textwidth]{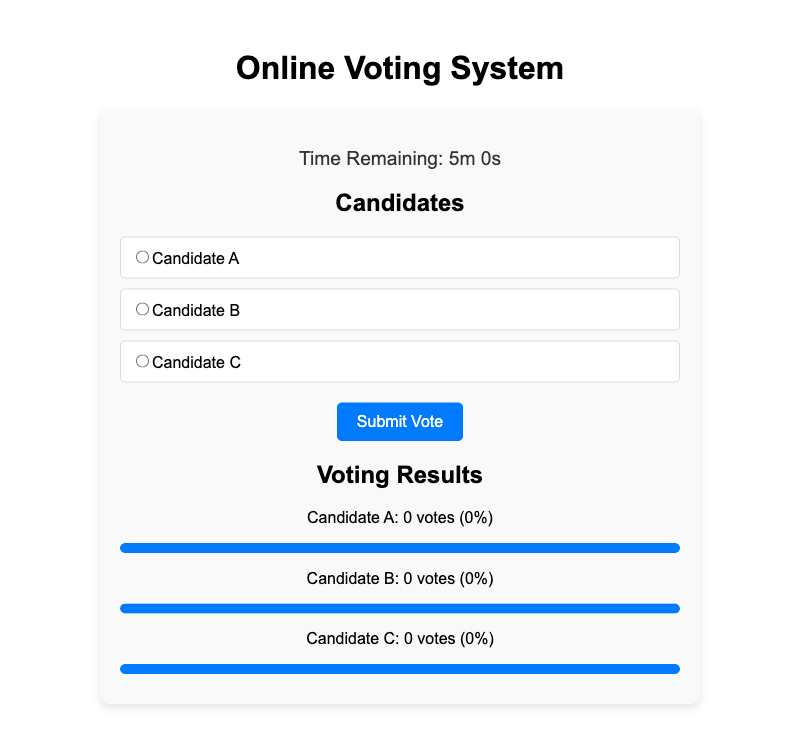}
        \label{fig:multi-img2}
    }
    \hfill
    \subfloat[Rendering result of the output code]{
        \includegraphics[width=0.3\textwidth]{multi_img2}
        \label{fig:multi_output}
    }
    \label{fig:multi_img_example}
\end{figure}

Previous implementation:
\begin{lstlisting}
import React, { useState } from 'react';

const OnlineVotingSystem = () => {
  const [candidates, setCandidates] = useState([{ name: 'candidate A', votes: 0 }, { name: 'candidate B', votes: 0 }]);
  const [selectedCandidate, setSelectedCandidate] = useState('');

  const handleVote = () => {
    setCandidates(candidates.map(candidate =>
      candidate.name === selectedCandidate ? { ...candidate, votes: candidate.votes + 1 } : candidate
    ));
  };

  return (
    <div className='container'>
      <div className='vote-list'>
        {candidates.map((candidate, index) => (
          <div key={index} className='vote-item'>
            <input type='radio' name='candidate' className='radio' onChange={() => setSelectedCandidate(candidate.name)} />
            {candidate.name}
          </div>
        ))}
      </div>
      <div className='result-section'>
        {candidates.map((candidate, index) => (
          <p key={index}>{candidate.name}\uff1a{candidate.votes} votes</p>
        ))}
      </div>
      <button className='vote-button' onClick={handleVote}>submit vote</button>
    </div>
  );
};

export default OnlineVotingSystem;
\end{lstlisting}

Generated code:
\begin{lstlisting}
import React, { useState, useEffect } from 'react';

const OnlineVotingSystem = () => {
  const [candidates, setCandidates] = useState([
    { name: 'Candidate A', votes: 0 },
    { name: 'Candidate B', votes: 0 },
    { name: 'Candidate C', votes: 0 },
  ]);
  const [selectedCandidate, setSelectedCandidate] = useState('');
  const [hasVoted, setHasVoted] = useState(false);
  const [showConfirmation, setShowConfirmation] = useState(false);
  const [errorMessage, setErrorMessage] = useState('');
  const [timeRemaining, setTimeRemaining] = useState(60 * 5);

  useEffect(() => {
    if (timeRemaining > 0) {
      const timer = setInterval(() => {
        setTimeRemaining((prev) => prev - 1);
      }, 1000);
      return () => clearInterval(timer);
    }
  }, [timeRemaining]);

  const handleVote = () => {
    if (!selectedCandidate) {
      setErrorMessage('Please select a candidate before voting.');
      return;
    }
    setShowConfirmation(true);
  };

  const confirmVote = () => {
    setCandidates((prevCandidates) =>
      prevCandidates.map((candidate) =>
        candidate.name === selectedCandidate
          ? { ...candidate, votes: candidate.votes + 1 }
          : candidate
      )
    );
    setHasVoted(true);
    setShowConfirmation(false);
    setErrorMessage('');
  };

  const cancelVote = () => {
    setShowConfirmation(false);
    setErrorMessage('');
  };

  const totalVotes = candidates.reduce((sum, candidate) => sum + candidate.votes, 0);

  return (
    <div className="voting-container">
      <h1>Online Voting System</h1>
      <div className="voting-content">
        <div className="time-remaining">
          <p>Time Remaining: {Math.floor(timeRemaining / 60)}m {timeRemaining % 60}s</p>
        </div>
        <div className="candidate-list">
          <h2>Candidates</h2>
          {candidates.map((candidate, index) => (
            <div key={index} className="candidate-item">
              <label>
                <input
                  type="radio"
                  name="candidate"
                  value={candidate.name}
                  onChange={() => setSelectedCandidate(candidate.name)}
                  disabled={hasVoted || timeRemaining <= 0}
                />
                {candidate.name}
              </label>
            </div>
          ))}
        </div>
        <button
          className="vote-button"
          onClick={handleVote}
          disabled={hasVoted || timeRemaining <= 0}
        >
          {hasVoted ? 'Voted' : 'Submit Vote'}
        </button>
        {errorMessage && <p className="error-message">{errorMessage}</p>}
        <div className="result-section">
          <h2>Voting Results</h2>
          {candidates.map((candidate, index) => (
            <div key={index} className="result-item">
              <p>
                {candidate.name}: {candidate.votes} votes (
                {totalVotes > 0 ? ((candidate.votes / totalVotes) * 100).toFixed(1) : 0}%)
              </p>
              <div className="progress-bar">
                <div
                  className="progress"
                  style={{ width: `${(candidate.votes / totalVotes) * 100}%` }}
                ></div>
              </div>
            </div>
          ))}
        </div>
        {showConfirmation && (
          <div className="confirmation-dialog">
            <p>Are you sure you want to vote for {selectedCandidate}?</p>
            <div className="dialog-buttons">
              <button onClick={confirmVote}>Yes</button>
              <button onClick={cancelVote}>No</button>
            </div>
          </div>
        )}
      </div>
    </div>
  );
};

export default OnlineVotingSystem;",
\end{lstlisting}

\subsection{Prompts For Data Synthesis}
In this section, we list all the prompts used in the processes of Evol-Instruct-Based synthesis, Waterfall-Model-Based synthesis, and Additive Development synthesis. 

\subsubsection{Prompts used in Evolution-Based Synthesis}
The prompt for code evolution is shown below:
\begin{lstlisting}
Objective: Enhance or modify the provided React code snippet, which includes both its JavaScript/TypeScript component code in js/jsx/ts/tsx format and associated CSS/SCSS/LESS styling. Aim to introduce meaningful changes that vary the original in terms of design, functionality, or structure, while ensuring the variation remains self-contained.

Self-contained Definition: The component must operate independently without external local resources such as additional CSS files, images, or third-party libraries. Styles and functionalities are entirely encapsulated within the provided code.

Guidelines for Generating Variations:

Breadth Directions (Diverse Modifications):

  - Visual Style Changes:
    - Modify color schemes, such as introducing a dark mode.
    - Adjust layout configurations, like grid versus flexbox settings.
    - Introduce transitions or animations for interactive elements.

  - Functional Extensions:
    - Add new functionalities like sorting, filtering, or pagination for data-driven components.
    - Integrate an API call to fetch data instead of static data use.
    - Add or remove components to adjust the overall structure, for example adding a button or a modal or something, any component that is defined in the package imported in the code, or a simple component that can be defined in the same file, in a word, DO NOT import any other package or component from other files to make the code self-contained!

  - Architectural Adjustments:
    - Split a large component into smaller, reusable sub-components.
    - Incorporate a higher-order component or custom hooks for state management or lifecycle methods.
    - Apply context API or Redux for more complex state management across the component tree.
    - Slightly add or remove features or components to adjust the overall structure, for example adding a button or a modal or something, any component that is defined in the package imported in the code, or a simple component that can be defined in the same file, in a word, DO NOT import any other package or component from other files to make the code self-contained!

Depth Directions (Complex Modifications):

  - Advanced State Management:
    - Use React's useReducer for managing complex component state instead of useState.
    - Implement custom state management logic that handles multiple states interdependently.

  - Performance Optimizations:
    - Employ techniques like lazy loading, suspense, or memoization to enhance performance.
    - Optimize conditional rendering to minimize the rendering operations.

  - Accessibility Enhancements:
    - Improve keyboard navigability and focus management within the component.
    - Ensure all interactive elements are accessible and ARIA-compliant, providing appropriate roles and properties.

  - Refactoring for Scalability:
    - Refactor the code to be more modular and maintainable, preparing it for scalability.
    - Ensure the component can handle varying levels of props or data volumes without degradation in performance.

Output Requirement: 
  - the output variation must not use the same variation strategy nor similar to the variations listed below
  - the output react code should also be self-contained.
  - do not repeat failed attempts listed below 
  - do not generate repeated code
  - the only allowed external resources are images with the same name as the original code, and the images are under the "imgs" folder, and more importantly, if using "import", the path should be "./imgs/<IMAGE_NAME>", if refering the image within the style, the path should be "url(/imgs/<IMAGE_NAME>)", if refering the image within the component code, the path should be "/imgs/<IMAGE_NAME>"
  - the output should only contain the react code and style code in the format "STYLE_VARIATION:<STYLE_VARIATION_CONTENT>
  ###COMPONENT_VARIATION:<COMPONENT_VARIATION_CONTENT>", no other comments or explainations or any other content is allowed in the output.

Variations:
\end{lstlisting}
The prompt for code similarity check is shown below:
\begin{lstlisting}
Objective: Determine if the newly generated variation of the React code snippet exhibits updates to previous variations or the original snippet, focusing on less critical changes in terms of updates, breadth, or depth.

Task: Analyze the newly generated React code snippet variation and compare it with previous variations and the original snippet. Evaluate changes in styling, functionality, or structural adjustments.

Criteria for Update Assessment:

- Style Updates: Examine if there are distinctive updates in the choice of colors, fonts, or layout modifications.
- Functional Enhancements: Check for updates in added functionalities, such as simple state changes, event handlers, or conditional rendering elements.
- Architectural Adjustments: Identify any resemblances in the way components are structured or organized, even if the changes are minor.
- Depth of Complexity: Look for similar complexity levels, even if the changes apply to less complex components or features.
- Breadth of Changes: Assess if there are minor correspondences in how various aspects of the snippet have been modified, focusing on subtler enhancements.

Output Requirement:

Respond with a single word: "Yes" or "No". This indicates whether the newly generated variation has distinctive updates in depth, or breadth when compared to any of the previous variations or the original.

Original Code snippet:
\end{lstlisting}

\subsubsection{Prompts Used in Waterfall-Model-Based Synthesis}
The prompt for system or application inference is shown below:
\begin{lstlisting}
Task: Given a code snippet, infer the system or application it belongs to. You should provide {infer_num} possible systems or applications that the code snippet could be part of, along with a one-sentence introduction of each system or application. You should use your full creativity and experience to come up with suitable systems or applications.

Output Format:
- List the {infer_num} possible systems or applications that the code snippet could be part of in bullet points.
- Each inferrence should be in large difference to others.
- After each system or application, provide a one-sentence introduction of the system or application.
- Each system or application together with its introduction SHOULD take only one line in the bullet point format.
- DO NOT include any additional commentary, explanation, or code blocks.
- The infered systems or applications can not be similar to the following examples:
{example_systems}

Code Snippet:
{code_snippet}
\end{lstlisting}

The prompt for requirement design inference is shown below:
\begin{lstlisting}
Task: Given a brief description of a system or application (single-page application). Infer the requirements of the system or application based on the common practices, design principles, and functionalities of similar systems or applications. You should first write a brief overview of the system or application, then list the requirements that often arise from such a system or application. Your expression should be in a clear, concise, and professional tone suitable for technical and stakeholder review.

Instructions and Output Format:
- List the inferred requirements as bullet points in a tone consistent with the product requirements document: detailed, clear, and professional.
- Do not include any additional commentary, explaination, or code blocks, only output the content with the following two parts (no other sections or headers are needed): 
  - A brief overview of the system or application.
  - A list of requirements that often arise from such a system or application based on the common practices, design principles, and functionalities of similar systems or applications.
  - You DO NOT need to raise the requirements in the follow perspective: Accessibility, Cross-Platform Compatibility, Security, Documentation and Support, Testing and Quality Assurance.
- The output should be formatted as: System Overview\n<SYSTEM_OVERVIEW_CONTENT>\n\nRequirements\n<REQUIREMENTS_LIST>

System Description: {system_description}
\end{lstlisting}

The prompt for designing requirements in the iterative process is shown below:
\begin{lstlisting}
Task: Review and make great modifications to the requirements of an existing system or application based on the implementation (including both the component code and the style code) from previous stage, real-world projects, real-life usage, and common industrial practices. 

Instructions and Output Format:
- Review Current Requirements: Begin by critically summarizing the existing project from the previous stage, highlight any functionalities, components, or styles that are currently implemented.
- Propose large modifications: Based on the summary, propose modificaitons to improve or modify the system's capabilities, functionalities, or design. Make sure the modifications are aligned with industrial standards and common practices. The modifications includes adding new features, modifying existing ones, or removing redundant or outdated components. The modifications should be large and significant (40 percent at least), not minor or trivial.
- The modificaitons should be formatted as a list of requirements in bullet points in a tone consistent with the product requirements document: detailed, clear, and professional.
- A list of requirements that often arise from such a system or application based on the common practices, design principles, and functionalities of similar systems or applications.
- You DO NOT need to raise the requirements in the follow perspective: Accessibility, Cross-Platform Compatibility, Security, Documentation and Support, Testing and Quality Assurance.
- Do not include any additional commentary, explaination, or code blocks, the output should be formatted as: Current Project Summary\n<CURRENT_PROJECT_SUMMARY_CONTENT>\n******\nProposed Modifications/Requirements\n<PROPOSED_MODIFICATIONS_LIST>

System description and requirements from previous stage:
{system_description}

Implementation from previous stage:
{code_snippet}
\end{lstlisting}

The prompt for layout design is shown below: 
\begin{lstlisting}
Task: Given a brief description of a system or application (single-page application) and its requirements, infer the layout of the system or application based on the common practices, design principles, and functionalities of similar systems or applications. You should list the layout components, their organization, and the interactions between them. Your expression should be in a clear, concise, and professional tone suitable for technical and stakeholder review.

Instructions and Output Format:
- List the inferred layout components, their organization, and interactions as bullet points in a tone consistent with the product requirements document: detailed, clear, and professional.
- Do not include any additional commentary, explaination, or code blocks, only the list (in bullet points) of layout components, their organization, and interactions based on the common practices, design principles, and functionalities of similar systems or applications.
- Do not wrap the output in any additional sections or headers or any markdown formatting.

System Description: 
{system_description}

Requirements: 
{requirements}
\end{lstlisting}

The prompt for layout design in the iterative process is shown below: 
\begin{lstlisting}
Task: Review and make great modifications to the layout of an existing system or application based on the layout description from previous stage, real-world projects, real-life usage, common industrial practices, and the provided requirements.

Instructions and Output Format:
- Modify the provided layout description based on the given requirements, you can add, modify, or remove layout components, their organization, and interactions. Make sure the modifications are aligned with industrial standards and common practices. The modifications should be large and significant (40 percent at least), not minor or trivial.
- The modified layout description should be aligned with the provided requirements.
- Do not include any additional commentary, explaination, or code blocks, only the list (in bullet points) of layout components, their organization, and interactions based on the common practices, design principles, and functionalities of similar systems or applications.
- Do not wrap the output in any additional sections or headers or any markdown formatting.

Layout description from previous stage:
{previous_layout_description}

Requirements:
{requirements}
\end{lstlisting}

The prompt for technical architecture design is shown below:
\begin{lstlisting}
Task: Given a brief description of a frontend React system or application (single-page application) and its requirements and layout, infer the technical architecture of the system or application based on the common practices, design principles, and functionalities of similar systems or applications. Remember this is a pure frontend system, and it is built using React. You should list the tech stack of the frontend development(which includes the libraries, frameworks (the framework has to be React), and tools used in the development of the system), the technical description of the functionalities, and the interactions between the components.
 Your expression should be in a clear, concise, and professional tone applicable for the engineering team to design and develop the system.

Instructions and Output Format:
- The output should consist of three parts: the tech stack, the technical description of the functionalities, and the interactions between the components.
- the tech stack is the libraries, frameworks (the framework has to be React), and tools used for the frontend development, do not include anything about the backend, or database or whatsoever.
- the tech stack should be primarily choosing the most common and widely used libraries and tools in the React ecosystem.
- the framework has to be React.
- the stack should include as least libraries as possible, only the most essential ones to implement the functionalities and design.
- You DO NOT need to include libraries like testing libraries, formatting libraries, babel, webpack, or any other libraries that are not directly related to the frontend development.
- The output should not contain any code, just the natural language description of the above three parts.
- Do not include any additional commentary, explaination, or code blocks, the three parts in the output should be in three separate paragraphs with headers.
- The output should be formatted as: 
  - Tech Stack\n<TECH_STACK_CONTENT>
  - Functionalities\n<FUNCTIONALITIES_CONTENT>
  - Interactions\n<INTERACTIONS_CONTENT>

System Description:
{system_description}

Requirements:
{requirements}

Layout:
{layouts}
\end{lstlisting}

The prompt for technical architecture design in the iterative process is shown below: 
\begin{lstlisting}
Task: Review and make great modifications to the technical architecture of an existing system or application based on the technical architecture description from the previous stage, real-world projects, real-life usage, common industrial practices, and the provided requirements and description of layout design.

Instructions and Output Format:
- The output should consist of three parts: the tech stack, the technical description of the functionalities, and the interactions between the components.
- the tech stack is the libraries, frameworks (the framework has to be React), and tools used for the frontend development, do not include anything about the backend, or database or whatsoever.
- the tech stack should be primarily choosing the most common and widely used libraries and tools in the React ecosystem.
- the framework has to be React.
- the stack should include as least libraries as possible, only the most essential ones to implement the functionalities and design.
- DO NOT use libraries or dependencies including: "react-i18next", "./redux/actions"
- You DO NOT need to include libraries like testing libraries, formatting libraries, babel, webpack, or any other libraries that are not directly related to the frontend development.
- The output should not contain any code, just the natural language description of the above three parts.
- The modifications should be aligned with the provided requirements and layout design.
- Do not include any additional commentary, explaination, or code blocks, the three parts in the output should be in three separate paragraphs with headers.
- The output should be formatted as: 
  - Tech Stack\n<TECH_STACK_CONTENT>
  - Functionalities\n<FUNCTIONALITIES_CONTENT>
  - Interactions\n<INTERACTIONS_CONTENT>

Technical Architecture from previous stage:
{previous_tech_architecture}

Requirements:
{requirements}

Layout:
{layouts}
\end{lstlisting}

The prompt for the design of the development plan is shown below: 
\begin{lstlisting}
Task: Design a step-by-step development plan for a frontend React single-page application based on the provided description, requirements, layout, and technical architecture. The development plan should focus exclusively on the coding and implementation of components and functionalities, not deployment, testing, or optimization tasks.

Instructions and Output Format:
- List 10 to 15 development tasks in the order they should be coded and completed.
- Each task must focus exclusively on coding-related tasks (e.g., developing components, integrating libraries, adding functionality). Do not include non-coding tasks like environment setup, testing, deployment, or optimization.
- The tasks must be designed for a single-page application and must not include multi-page functionality.
- Each task must specify what to implement, how to implement it, and any specific libraries or tools to use, written in a concise, clear, and professional tone.
- The output must strictly follow the format below:
  - Task 1  
  <TASK_1_CONTENT>  

  - Task 2  
  <TASK_2_CONTENT>  

  - Task 3  
  <TASK_3_CONTENT>  
  ...  

Do not include any additional commentary, explanations, or code blocks outside the task descriptions.

Example Output Template:
  - Task 1  
  Implement the header component. Use React functional components and include a navigation bar with placeholders for links. Style the component using styled-components.  

  - Task 2  
  Develop the footer component. Include contact information and social media icons. Use FontAwesome for icons and ensure responsive design using CSS Grid.  

System Description:
{system_description}

Requirements:
{requirements}

Layout:
{layouts}

Technical Architecture:
{tech_architecture}
\end{lstlisting}

The prompt for code generation is shown below: 
\begin{lstlisting}
Task: Given an implementation of a frontend React system or application (single-page application) with a brief system introduction, additively update the current implementation according to the given task description. The code snippet should be consistent with the common development practices and React component design principles used in real-world projects. The code snippet should be self-contained and consistent with the previous code snippets in the development plan.

Instructions:
- The code snippet should be additively developed upon the Current Implementation (If there the current implementation is not <EMPTY>, otherwise, you need to implement the task based on the given code, and DO NOT delete any part of the Current Implementation if it can introduce not conflict with the given task description). 
- You MUST implement exactly the functionalities, layout, together with any details described in the task description.
- The code snippet must operate independently without any external local resources such as additional local files, images, or data. 
- The functionalities are entirely encapsulated within the provided code.
- If the code snippet requires data or any kind of input, it should be hard-coded within the component (if input is required, there should be default values for the input).
- The code snippet should be in JavaScript or TypeScript for the component code, and CSS for the styling.
- The output code snippet should be a complete component that can be rendered in a React application, including the import statements, component definitions, export statements, styling, and any other necessary code like event handlers, state management, or mock data for the component to function.
- DO NOT wrap the output and code blocks in any additional sections or headers or any markdown formatting.
- DO NOT generate repeated code.
- The code snippet must have one single default export component (The most top-level component).
- DO NOT use packages or depencies including: "react-i18next", "./redux/actions"
- The code style must be aligned with the common React development practices in real-world projects, for example using components, hooks, or anything according to your knowledge as an expert frontend engineer.
- The code snippet should not include any comments, explanations, or additional content. ONLY the code snippet in the format "STYLE:<STYLE_CONTENT>###COMPONENT:<COMPONENT_CONTENT>".

System Introduction:
{system_description}

Current Implementation:
{current_implementation}

Task Description:
{next_task_description}
\end{lstlisting}

The prompt for code check is shown below: 
\begin{lstlisting}
Review the given code. Ensure the following requirements are met:

1. Self-Containment
- Ensure that the generated component operates independently without relying on external resources such as files, images, or external data (e.g., API calls). If the component requires any input data, it should be hardcoded (mocked) within the component itself, using default values wherever necessary.
- All dependencies must be included directly in the code (i.e., there should be no missing imports or external files).
2. Code Structure and Format
- The generated code should not include any additional sections, headers, or markdown formatting. It must follow this format:
  "STYLE:<STYLE_CONTENT>###COMPONENT:<COMPONENT_CONTENT>"
- The code should include:
  - A single default export component that is the top-level component.
  - Proper imports, including React and necessary utilities.
  - All necessary event handlers, state management, and any mock data.
  - No additional comments or explanations in the code.
3. Avoid Redundancies
- Ensure that there is no repeated or redundant code. Each function, variable, and component should be used only once unless necessary for the design.

Input code:
{code_snippet}

If the code meets all the requirements, respond with a single word "Passed." If there are any issues or violations, make the necessary corrections and provide the updated code in the format "STYLE:<STYLE_CONTENT>###COMPONENT:<COMPONENT_CONTENT>". NO additional comments or explanations are needed.
\end{lstlisting}

\subsubsection{Prompts Used in Additive Development Synthesis}
The prompt for system or application inference is shown below: 
\begin{lstlisting}
Given a React code snippet that I want to integrate into a larger, real-world, production-grade, single-page frontend system. Your task is to propose exactly {infer_num} distinct, fully-featured production-ready frontend React systems where this code snippet can play a meaningful role. Each system should be self-contained, modular, and entirely client-side with no dependencies on backend APIs or real-time data sources.

Each proposed system must be a real, useful product or tool that could exist in a commercial, industrial, or enterprise context. The system should feel like something that could be released as a standalone product (like a SaaS tool, utility app, or data visualization tool) and not a "demo" project.

Proposal Requirements
1. System Name: Provide a name for the system.
2. System Category/Type: Identify the type of system (e.g., interactive dashboard, design tool, productivity app, data visualization app, etc.).
3. System Purpose and Use Case: Describe the purpose of the system, its primary function, and what real-world problem it solves.
4. How the Provided Code Snippet Fits: Clearly explain how and where the provided React code snippet would be used as a key part of the system (e.g., as a core component, widget, or logic handler).
5. System Complexity: Each system must include at least 3-4 interconnected components or sub-modules (not pages) that logically interact with each other.
6. Core Features: Each system should have at least 5-7 essential production-grade features. Example features include:
  - Data visualization and graphs (e.g., charts, dashboards)
  - Interactive forms and filters (e.g., dynamic search, multi-step form validation)
  - Dynamic UI updates (without real-time data - instead, use hardcoded, local JSON or JavaScript objects as data)
  - Stateful UI logic (e.g., tabs, modals, tooltips, or collapsible views)
  - Interactive elements (e.g., drag-and-drop, sliders, sortable lists, or resizable panels)
  - Responsive design (ensure responsiveness across desktop, tablet, and mobile)

Important Guidelines:
- The systems must be built as self-contained, single-page applications. Avoid multiple pages, page reloads, or navigation logic (like "404 pages" or "about pages").
- The system must feel like a standalone product that could be used by a business or an individual. It should not be a "demo" project.
- Each system should use mocked local data (e.g., hard-coded JSON, JavaScript arrays, or default props) for any data-driven features. No real-time data or backend API calls should be used.
- Each system should have distinct logic, purpose, and design. For example, a "To-Do List" app and a "Task Planner" are too similar.
- The output MUST only contain the proposed systems in the following JSON format (you MUST NOT change the key names or add any additional content):
  [
    {{
      "name": "System Name",
      "category": "System Category",
      "purpose": "System Purpose and Use Case",
      "code_snippet_usage": "How the Provided Code Snippet Fits",
      "complexity": "System Complexity",
      "features": "Core Features"
    }},
    {{
      "name": "System Name",
      "category": "System Category",
      "purpose": "System Purpose and Use Case",
      "code_snippet_usage": "How the Provided Code Snippet Fits",
      "complexity": "System Complexity",
      "features": "Core Features"
    }},
    ...
  ]
  NO other comments, explainations, or any content is needed, and do not wrap with markdown.

- The infered systems or applications can not be similar to the following examples:
{example_systems}

Code Snippet:
{code_snippet}
\end{lstlisting}

The prompt for the design of development plan is shown below: 
\begin{lstlisting}
Given a description of a pure frontend system/application, create a complete, step-by-step development roadmap for building this system from start to finish. The development process should follow an additive approach, meaning each step builds on the previous one, introducing new logic, components, and interactive features. The goal is to create a fully-functional, production-grade, standalone React application.

Development Plan Requirements
1. General Structure
  - The system should be developed over at least 15-20 development steps to ensure sufficient complexity and production-grade features.
  - Each step should be large enough to feel like a deliverable milestone, and there MUST be some changes on the visual outlook as well(for example adding/removing/updating components or updating layouts). Each step should introduce significant system-level features, logic updates, or interactive functionality.
  - Each step must be self-contained and should culminate in a single, large, production-grade, single-file React application. This means that all components, styles, logic, and state must be written in one large self-contained code chunk (no imports from local files).
  - The updated should avoid invisible changes (e.g., only updating comments, minor code formatting, or the components can only be seen if some interactions are made).
  - No external data sources or real-time data can be used. If data is needed, it must be hardcoded in the component logic using JSON objects, JavaScript arrays, or mock data.
  - No file separation is allowed. All components, logic, and functionality must exist within a single large code block (one big self-contained React file).

2. Details for Each Step
  - Step Title: A clear name for the development task (e.g., "Build a Filterable Table", "Create a Dashboard with Sorting and Pagination").
  - Objective: The purpose of this step (e.g., "Enable users to sort the table columns dynamically").
  - Components/Logic Introduced: Specify which new logic, components, or features are introduced in this step.
  - How It Builds on the Previous Step: Explain how this step logically builds on the previous one (e.g., "Now that data is displayed, this step enables interactive filtering").
  - Best Practices: Indicate best practices being followed (e.g., "Use memoization to optimize render performance", "Use DRY principles to avoid repeating logic").

3. Example Development Process (3 steps for illustration)
  - Step 1
    Set Up Initial Layout and Component Structure
    Objective: Create the initial layout and component structure for the dashboard.
    Components/Logic Introduced:
      Create Header, Sidebar, and Main Content Area as self-contained elements in the code.
      Hardcode sample navigation items in the sidebar (like "Home", "Reports", "Settings").
      Create an empty Data Display Area where dynamic components (like charts or tables) will be rendered later.
    How It Builds on the Previous Step: Since this is the first step, it sets the foundation for later steps where logic, interactivity, and dynamic features will be added.
    Best Practices:
      Use reusable functions and self-contained components.
      Use Flexbox or CSS Grid to create a responsive layout.
  - Step 2
    Create a Data Table with Hardcoded Data
    Objective: Create a simple table component that displays hardcoded data.
    Components/Logic Introduced:
      Add a DataTable inside the Main Content Area.
      Use map() to iterate over hardcoded data and render each row dynamically.
      Add simple headers (like "Name", "Age", "Role", "Location").
    How It Builds on the Previous Step: The DataTable is displayed inside the Main Content Area created in Step 1.
    Best Practices:
      Pass data as a variable at the top of the file, not as an external file import.
      Use array map() to generate table rows dynamically.
  - Step 3
    Add Sorting Functionality to the Data Table
    Objective: Allow users to sort table columns by clicking on the headers (e.g., clicking "Age" sorts the table by age).
    Components/Logic Introduced:
      Add sort state to track the column being sorted and the sort order (ascending/descending).
      Modify the table headers so that clicking on them triggers the sort.
    How It Builds on the Previous Step: Builds on the existing DataTable by adding interactivity.
    Best Practices:
      Use React state to track sorting.
      Optimize sorting logic with React.memo.

4. Final Requirements for the Development Plan
  - Self-Contained Single-File Code: The final system, when fully implemented, should exist in a single, large React file. All components, logic, and styles must exist within this file. No imports of local files, CSS files, or additional components are allowed.
  - Hardcoded Data Only: If any data is required for the system, it must be stored directly in the file using hardcoded objects, arrays, or default values.
  - Single-Page Application (SPA): The system must not use any page-based navigation logic. All interactions should take place within the same page.
  - Fully Incremental Steps: The system should evolve naturally as each development task is completed, with each task adding significant logic or interactivity.
  - Complexity and Depth: Ensure the system has sufficient depth and complexity. It should have at least 15-20 development steps to demonstrate meaningful growth and progression.
  - The output MUST only contains the step list in the JSON format (you MUST NOT change the key names or add any additional content): 
    [
      {{
        "title": "Step Title",
        "objective": "Objective Description",
        "components_logic": "Components/Logic Introduced",
        "builds_on": "How It Builds on the Previous Step",
        "best_practices": "Best Practices Followed"
      }},
      {{
        "title": "Step Title",
        "objective": "Objective Description",
        "components_logic": "Components/Logic Introduced",
        "builds_on": "How It Builds on the Previous Step",
        "best_practices": "Best Practices Followed"
      }},
      ...
    ]
  no titles, headings, comments or any other content is needed, and do not wrap with markdown.

System Description:
{system_description}

Code Snippet:
{code_snippet}
\end{lstlisting}

The prompt for code generation is shown below: 
\begin{lstlisting}
Task: Given an implementation of a frontend React system or application (single-page application) with a brief system introduction, current development task and current implementation of the system. Additively update the current implementation according to the given task description. The generated code should be consistent with the common development practices and React component design principles used in real-world projects. The generated code should be self-contained. You should also learn the coding style and design patterns from the reference code as well, incorporate the similar implementation if the reference code matches the current task.

Instructions:
- The generated code should be additively developed upon the Current Implementation (you need to implement the task based on the Current Implementation, you MUST NOT start from scratch). 
- You MUST implement exactly the functionalities, layout, together with any details described in the task description.
- The generated code must operate independently without any external local resources such as additional local files, images, or data. 
- The functionalities are entirely encapsulated within the generated code.
- If the generated code requires data or any kind of input, it should be hard-coded within the component (if input is required, there should be default values for the input).
- The generated code should be in JavaScript or TypeScript for the component code, and CSS for the styling.
- The output generated code should be a complete component that can be rendered in a React application, including the import statements, component definitions, export statements, styling, and any other necessary code like event handlers, state management, or mock data for the component to function.
- DO NOT wrap the output and code blocks in any additional sections or headers or any markdown formatting.
- DO NOT generate repeated code.
- The generated code must have one single default export component (The most top-level component).
- DO NOT use packages or depencies including: "react-i18next", "./redux/actions"
- The code style must be aligned with the common React development practices in real-world projects, for example using components, hooks, or anything according to your knowledge as an expert frontend engineer.
- The style spcification must be implemented with the styled-components and no other CSS, SCSS, or LESS specifications are allowed.
- The generated code should not include any comments, explanations, or additional content. ONLY the generated code.

System Introduction:
{system_description}

Current Implementation:
{current_implementation}

Task Description:
{next_task_description}
\end{lstlisting} 

The prompt for code check is shown below: 
\begin{lstlisting}
Review the given code. Ensure the following requirements are met:

1. Self-Containment
- Ensure that the generated component operates independently without relying on external resources such as files, images, or external data (e.g., API calls). If the component requires any input data, it should be hardcoded (mocked) within the component itself, using default values wherever necessary.
- All dependencies must be included directly in the code (i.e., there should be no missing imports or external files).
2. Code Structure and Format
- The generated code should not include any additional sections, headers, or markdown formatting.
- The code should include:
  - A single default export component that is the top-level component.
  - Proper imports, including React and necessary utilities.
  - All necessary event handlers, state management, and any mock data.
  - No additional comments or explanations in the code.
3. Avoid Redundancies
- Ensure that there is no repeated or redundant code. Each function, variable, and component should be used only once unless necessary for the design.

Input code:
{code_snippet}

If the code meets all the requirements, respond with a single word "Passed." If there are any issues or violations, make the necessary corrections and answer with the updated code ONLY. NO additional comments or explanations are needed.
\end{lstlisting}

\subsection{Prompts for Training}\label{apdix:train-prompt}
The prompt used in the middle stage of training, in which the vision encoder and connector are trained while keeping the LLM frozen, is shown below.
\begin{lstlisting}
Analyze the provided image of a webpage screenshot. Infer the task description that would be required to create this image, and describe the observed layout of its components.
### Input Image:
{image}

### Response:

\end{lstlisting} 

The prompt for the task of "Interpretation Before Coding" is shown below:
\begin{lstlisting}
Below is an image of the page we want to create. Generate React code to replicate the design, including layout, typography, and styling. Also provide a layout description. Format your response as follows:
'// Layout Description\n[Describe component layout]\n\n// CSS\n[CSS/SCSS code]\n\n// [React Implementation (JS/TS/JSX/TSX)]\n[Component code]'.

### Input Image:
{image}

### Response:  
\end{lstlisting}

The prompt for the task of "Code Only" is shown below:
\begin{lstlisting}
Below is an image of the page to create. Generate React code and styles to replicate the design, including layout, typography, and styling. Format your response as follows:'// CSS\n[CSS/SCSS code]\n\n// [React Implementation (JS/TS/JSX/TSX)]\n[Component code]'.

### Input Image:
{image}

### Response:  
\end{lstlisting}

The prompt for the multi-image scenario is shown below: 
\begin{lstlisting}
Analyze two provided images: one showing a webpage (image_1) and another showing its modified version (image_2). Identify visual differences, such as layout, content, or style changes. Based on the provided code snippet for image_1, generate an updated code snippet that reflects the modifications seen in image_2.
Format your response as follows:'// CSS\n[CSS/SCSS code]\n\n// [React Implementation (JS/TS/JSX/TSX)]\n[Component code]'.

### image_1:
{image_1}

image_2:
{image_2}

###code snippet for image_1: 
{css_code_for_image_1}

{imp_code_for_image_1}

### Response:
\end{lstlisting}

\subsection{Illustration of Image Matching Based on Similarity Scores}\label{apdx:sim-show}
We randomly select a test case and present the corresponding rendered images from the generated code, sorted by similarity scores, in Table \ref{tab:image-sim-table}. 
The first image serves as the reference, while the following nine rendered images are displayed with their respective similarity scores labeled. 
It can be observed that images with similarity scores above 0.9 closely align with human perception.

\begin{table*}[htbp]
  \renewcommand{\arraystretch}{1.5}
  \centering
  \begin{tabular}{cccc}
      \fbox{\begin{minipage}{0.32\textwidth}
          \centering
          \includegraphics[width=\linewidth]{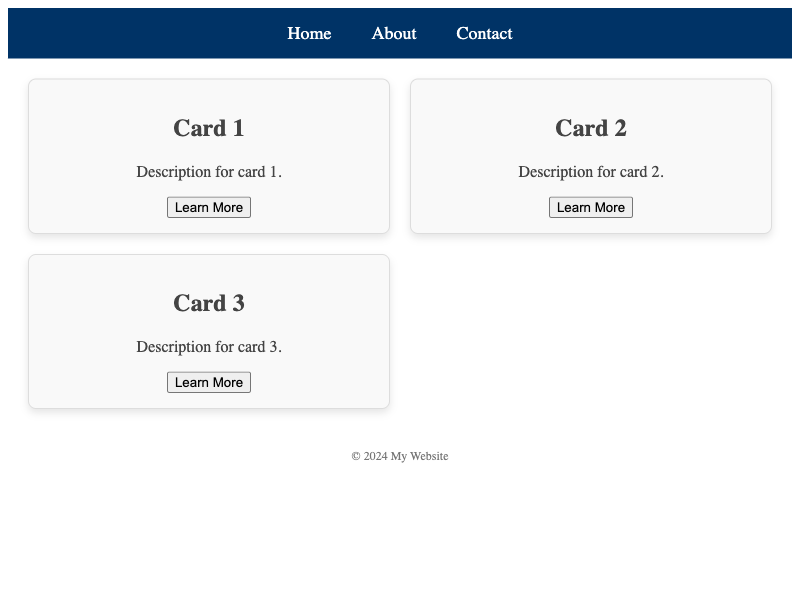}
          \caption*{\centering screenshot of a test case }
      \end{minipage}} &
      \begin{minipage}{0.28\textwidth}
          \centering
          \includegraphics[width=\linewidth]{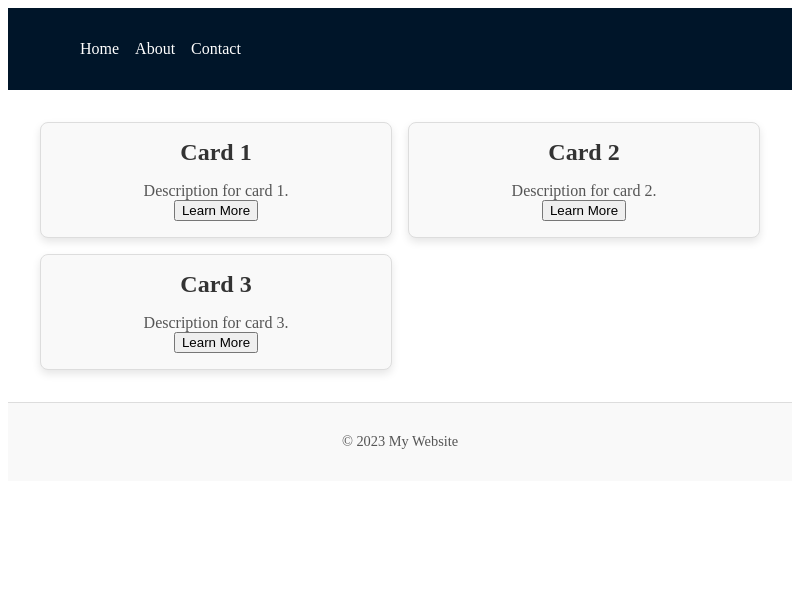}
          \caption*{0.975}
      \end{minipage} &
      \begin{minipage}{0.28\textwidth}
          \centering
          \includegraphics[width=\linewidth]{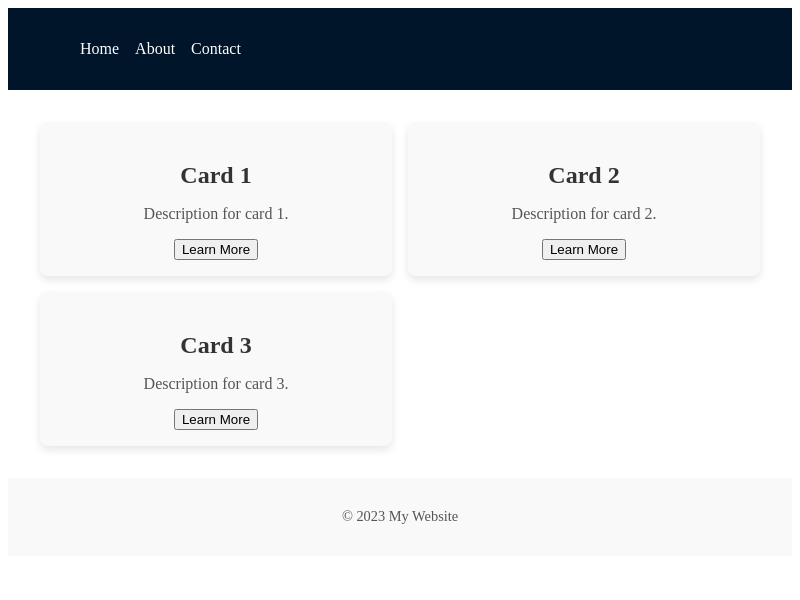}
          \caption*{0.967}
      \end{minipage}  \\ \rule{0pt}{20ex}

      \begin{minipage}{0.28\textwidth}
          \centering
          \includegraphics[width=\linewidth]{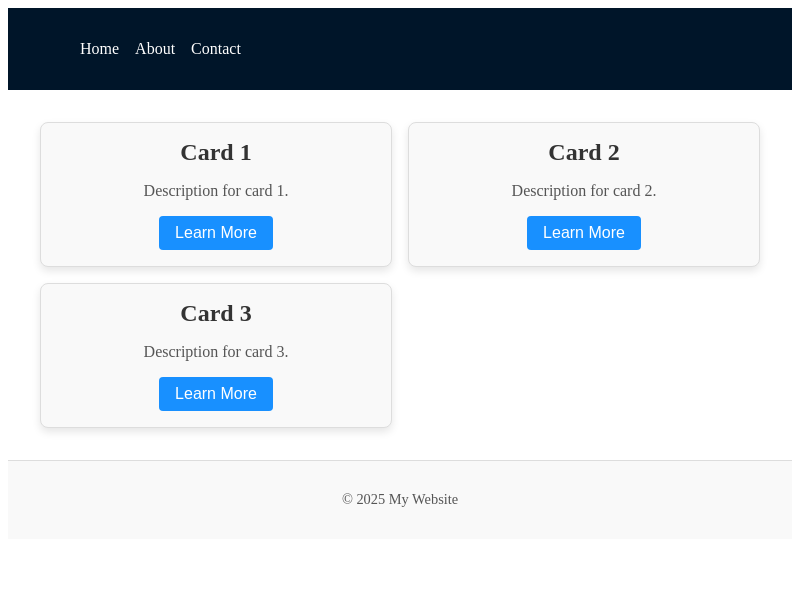}
          \caption*{0.966}
      \end{minipage} &
      \begin{minipage}{0.28\textwidth}
          \centering
          \includegraphics[width=\linewidth]{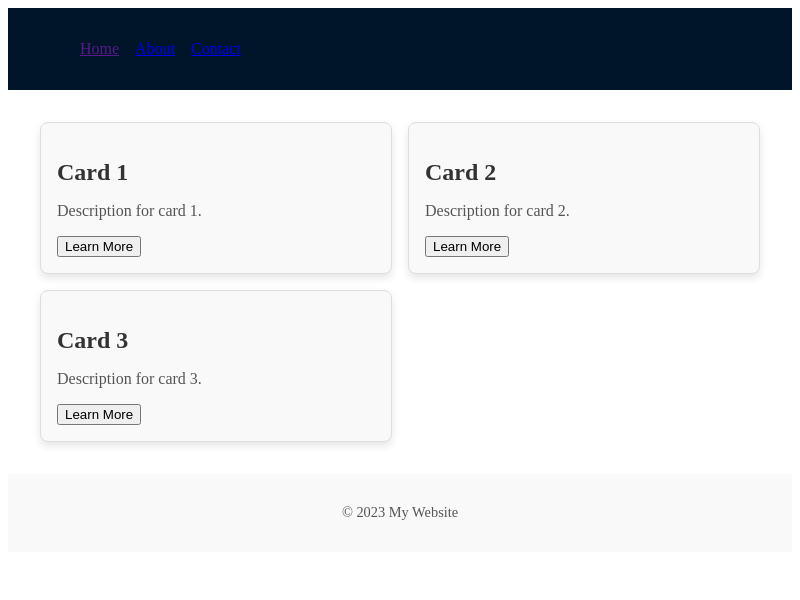}
          \caption*{0.960}
      \end{minipage} &
      \begin{minipage}{0.28\textwidth}
          \centering
          \includegraphics[width=\linewidth]{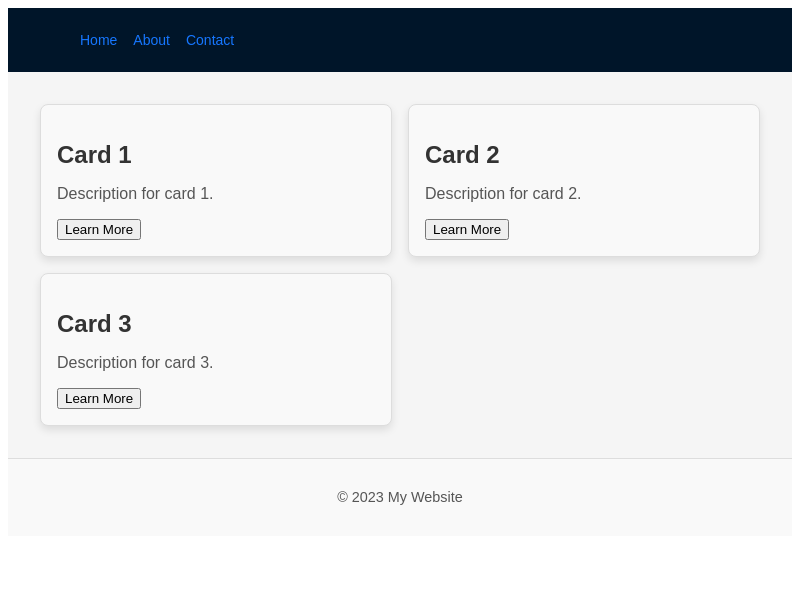}
          \caption*{0.954}
      \end{minipage} \\ \rule{0pt}{20ex}

      \begin{minipage}{0.28\textwidth}
          \centering
          \includegraphics[width=\linewidth]{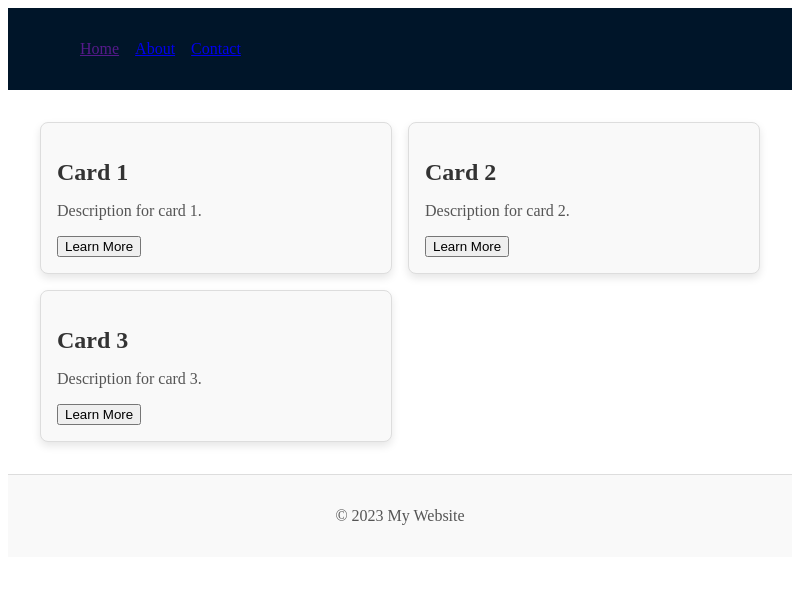}
          \caption*{0.939}
      \end{minipage} &
      \begin{minipage}{0.28\textwidth}
          \centering
          \includegraphics[width=\linewidth]{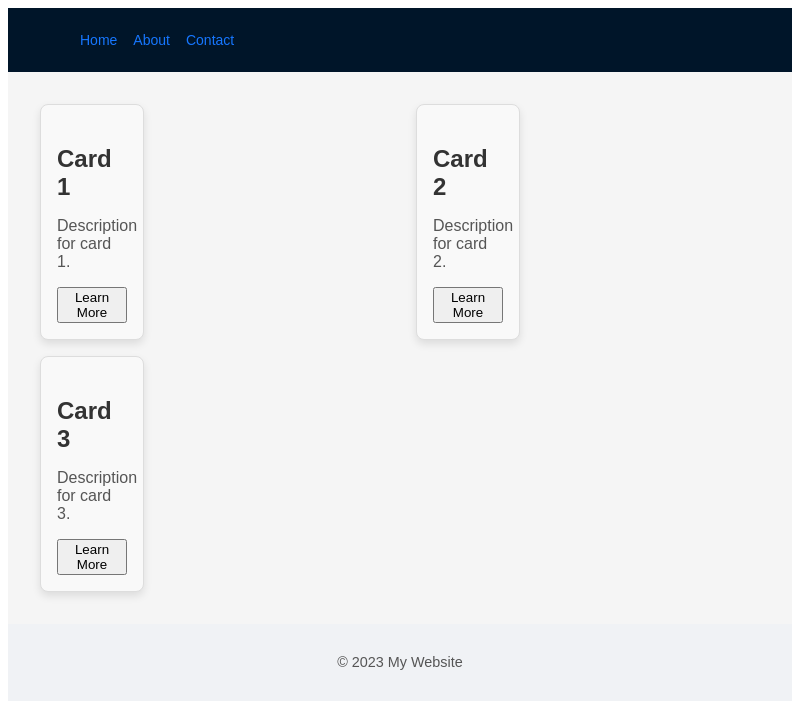}
          \caption*{0.880}
      \end{minipage} &
      \begin{minipage}{0.28\textwidth}
          \centering
          \includegraphics[width=\linewidth]{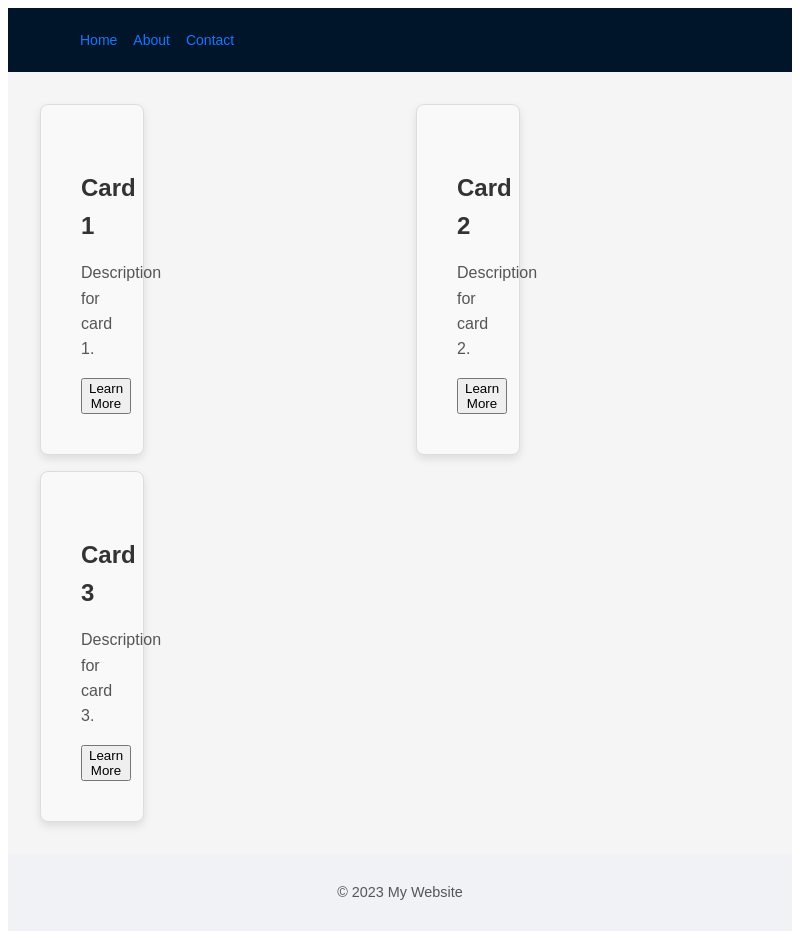}
          \caption*{0.775}
      \end{minipage} &
  \end{tabular}
  \caption{Screenshots ranked by similarity to the test case (Top-Left), based on cosine similarity of embeddings computed by DINOv2-Base.}
  \label{tab:image-sim-table}
\end{table*}

\end{document}